\documentclass[10pt,twocolumn,letterpaper]{article}

\usepackage{cvpr}              % To produce the CAMERA-READY version
\usepackage{algorithm}      % 提供 algorithm 浮动体环境
\usepackage{algorithmic}    % 提供 algorithmic 环境和命令（如 \REQUIRE、\ENSURE）version
\usepackage{booktabs}    % 提供 \toprule, \midrule, \bottomrule 等高质量表格线
\usepackage{adjustbox}   % 提供 \adjustbox{} 命令，用于调整表格整体宽度
\usepackage{amssymb}     % 提供数学符号，如 \xmark, \checkmark
\usepackage{graphicx}    % 提供图形和缩放支持（adjustbox 内部也依赖）
\usepackage{multirow} 
\usepackage{pifont}
\usepackage[table]{xcolor}
\newcommand{\xmark}{\ding{55}}  % 定义叉号符号
\definecolor{cvprblue}{rgb}{0.21,0.49,0.74}
\usepackage[pagebackref,breaklinks,colorlinks,allcolors=cvprblue]{hyperref}

\def\paperID{9367} % *** Enter the Paper ID here
\def\confName{CVPR}
\def\confYear{2026}

\title{ When Models Learn to Ask Why: Adaptive Causal Reasoning for Trustworthy Medical Vision–Language Models}
\author{
Jianxin Lin$^{1,*}$, 
Chunzheng Zhu$^{2,*}$, 
Peter J. Kneuertz$^{1}$, 
Yunfei Bai$^{3,\ddagger}$, 
Yuan Xue$^{1,\dagger}$\\[-0.2em]
$^{1}$The Ohio State University 
$^{2}$Hunan University 
$^{3}$Amazon\\
{\tt\small xue.643@osu.edu}
}

\begin{document}
\maketitle
\begingroup
\renewcommand\thefootnote{}\footnotetext{* Equal contribution. $\dagger$ Corresponding author. $\ddagger$ Work conducted in an individual capacity and does not reflect the views of Amazon.}
\addtocounter{footnote}{-1}
\endgroup
\begin{abstract}

Vision-Language Models (VLMs) have enabled interpretable medical diagnosis by integrating visual perception with linguistic reasoning. Yet, existing medical chain-of-thought (CoT) models lack explicit mechanisms to represent and enforce causal reasoning, leaving them vulnerable to spurious correlations and limiting their clinical reliability. We pinpoint three core challenges in medical CoT reasoning: how to adaptively trigger causal correction, construct high-quality causal-spurious contrastive samples, and maintain causal consistency across reasoning trajectories. To address these challenges, we propose MedCausalX, an end-to-end framework explicitly models causal reasoning chains in medical VLMs. We first introduce the CRMed dataset providing fine-grained anatomical annotations, structured causal reasoning chains, and counterfactual variants that guide the learning of causal relationships beyond superficial correlations. Building upon CRMed, MedCausalX employs a two-stage adaptive reflection architecture equipped with \texttt{<causal>} and \texttt{<verify>} tokens, enabling the model to autonomously determine when and how to perform causal analysis and verification. Finally, a trajectory-level causal correction objective optimized through error-attributed reinforcement learning refines the reasoning chain, allowing the model to distinguish genuine causal dependencies from shortcut associations. Extensive experiments on multiple benchmarks show that MedCausalX consistently outperforms state-of-the-art methods, improving diagnostic consistency by +5.4 points, reducing hallucination by over 10 points, and attaining top spatial grounding IoU, thereby setting a new standard for causally grounded medical reasoning. The code and dataset are available at \url{https://github.com/zhcz328/MedCausalX}.

\end{abstract}

\begin{figure}
    \centering
    \includegraphics[width=1\linewidth]{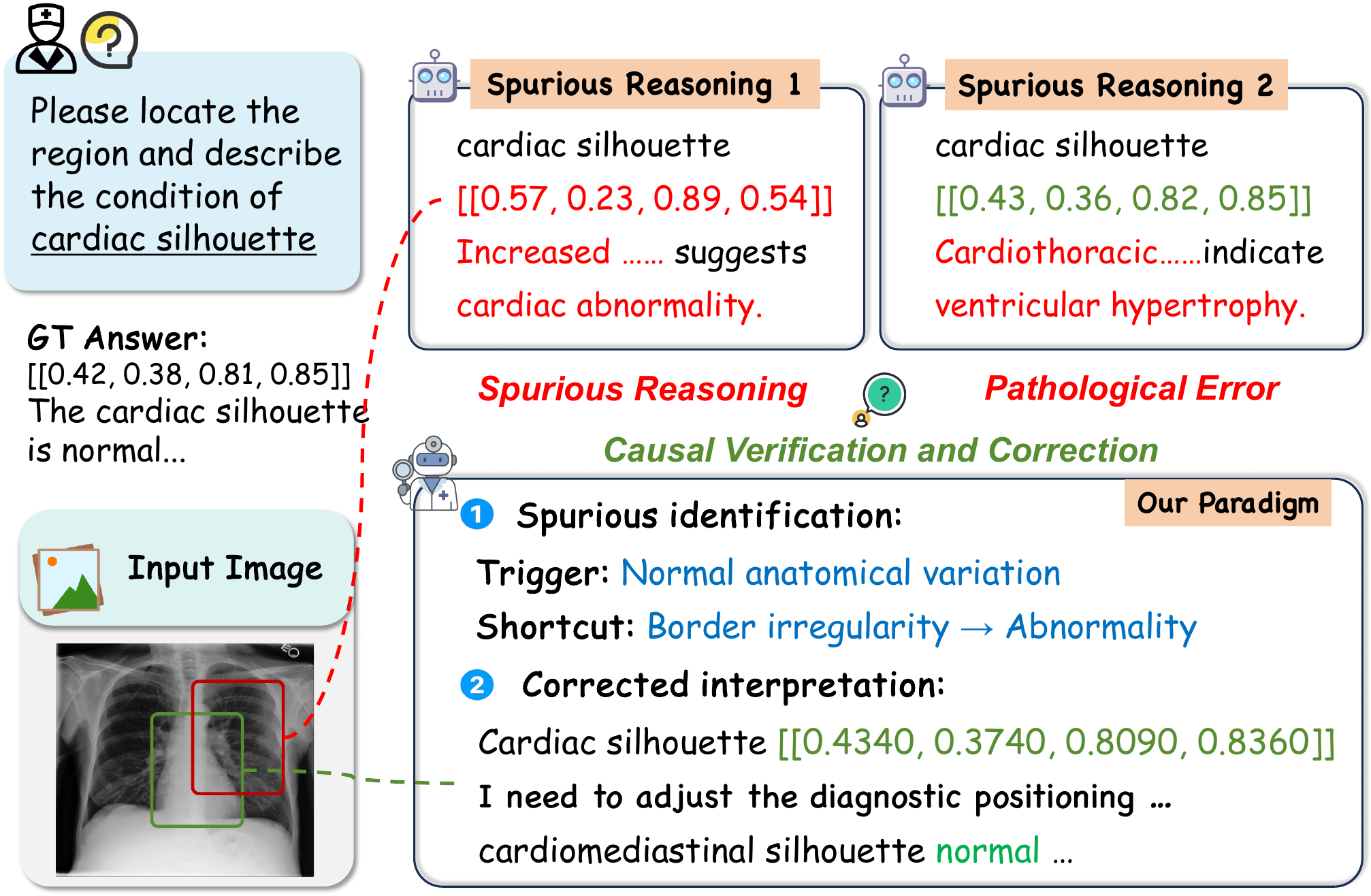}
    \caption{Existing medical VLMs often lack causality-awareness, which manifests as \textit{(1)} spurious CoT reasoning attends to mislocalized anatomy or \textit{(2)} overlooking causal evidence for pathology diagnosis. MedCausalX can achieve trustworthy diagnosis by adaptive causal verification and correction during inference.}
    \label{fig:placeholder}
\end{figure}
\section{Introduction}

Vision-Language Models (VLMs) have shown remarkable potential in medical imaging by coupling visual perception with natural language reasoning for tasks such as diagnostic interpretation and report generation~\cite{achiam2023gpt, deng2024enhancing,huang2025medreflect,gong2025med}.
% By processing large-scale multimodal data, these models can provide intelligent clinical decision support~\cite{tanno2025collaboration,rao2025multimodal}.
By processing large-scale multimodal data, these models can provide 
intelligent clinical decision support~\cite{tanno2025collaboration,
rao2025multimodal,chenbridging,chuang2025tinymig,liu2025perturbating}.
To enhance the reasoning transparency, chain-of-thought (CoT) prompting has been integrated into medical VLMs~\cite{pan2025medvlm,lai2026med,liu2024medcot}, producing intermediate steps that make the diagnostic process interpretable.

Despite these advances, current medical CoT models lack explicit mechanisms to represent and enforce causal dependencies, which limits their robustness and clinical reliability~\cite{geirhos2020shortcut,lin2024shortcut,liu2024intrinsic}.
As a result, models may implicitly exploit superficial correlations, such as global background features, language model priors, or training distribution biases~\cite{xie2025training}, rather than capturing true causal relations between anatomy, pathology, and diagnosis. This gap between pattern recognition and causal reasoning can lead to explanations that sound plausible yet are not fully aligned with the underlying clinical mechanisms.

In clinical practice, causal reasoning underpins diagnostic reliability~\cite{richens2020improving,castro2020causality,peters2017elements}. Clinicians reason through a structured chain: identify anatomical abnormalities, interpret pathological manifestations, and draw diagnostic conclusions grounded in established medical knowledge~\cite{vlontzos2025causal,prosperi2020causal}. This process ensures that decisions are supported by causal mechanisms rather than surface correlations, a property current medical VLMs do not yet reliably exhibit.

% Recent progress in synthetic data construction~\cite{li2025retidiff} and reinforcement learning has enabled the modeling of structured reasoning trajectories~\cite{zhang2024improve,shao2024deepseekmath,pan2025medvlm,lai2026med}, offering new opportunities for causal training in medical domains. However, learning effective dynamic CoT reasoning faces three open challenges.
Despite recent progress in structured reasoning trajectories~\cite{zhang2024improve,shao2024deepseekmath,pan2025medvlm,lai2026med}, learning effective dynamic CoT reasoning faces three open challenges.
First, how to dynamically identify when causal correction is needed during reasoning rather than statically applying CoT, which can entrench incorrect causal associations. Second, how to construct high-quality contrastive samples that juxtapose causal and non-causal associations to facilitate adaptive reasoning verification while minimizing expert annotation requirements. Third, how to ensure end-to-end causal consistency across the entire reasoning trajectory instead of optimizing token-level likelihoods that ignore global causal coherence.

To systematically address these challenges, we propose MedCausalX, an end-to-end framework that explicitly models causal reasoning with adaptive reflection in medical VLMs.
MedCausalX is grounded in three causal insights:
\textit{(1)} Causal compositionality: diagnosis unfolds as a structured causal hierarchy where anatomical localization conditions pathological characterization, and both jointly determine diagnostic conclusions;
\textit{(2)} Causal disruption detection: reasoning failures often arise from local violations of causal dependencies, motivating adaptive reflection to identify and correct these disruptions; and
\textit{(3)} Trajectory-level causal alignment: enforcing global consistency across the reasoning trajectory, beyond token-level likelihood, aligns the model’s reasoning process with causal semantics and discourages reliance on superficial correlations.
%
%Guided by these principles, MedCausalX establishes a unified framework that integrates dataset-level causal supervision, adaptive reasoning mechanisms, and hierarchical causal optimization into a coherent learning paradigm.
%It builds upon the CRMed dataset, which provides fine-grained anatomical annotations, structured causal reasoning chains, and counterfactual variants that enable both causal supervision and robustness evaluation.
%To achieve adaptive reasoning, MedCausalX employs a two-stage reflection mechanism with \texttt{<causal>} and \texttt{<verify>} tokens, allowing the model to transition from causal analysis to verification and correction in a self-directed manner.
%Finally, a hierarchical causal optimization pipeline sequentially integrates trajectory-level supervised learning with reinforcement learning on error-attributed trajectories, enforcing causal consistency across the reasoning process and disentangling genuine causal dependencies from shortcut associations.

Guided by these principles, MedCausalX establishes a unified framework integrating data-level causal supervision, an adaptive reasoning mechanism, and hierarchical optimization into a coherent learning paradigm. It builds upon the CRMed dataset, which provides fine-grained anatomical annotations, structured causal reasoning chains, and counterfactual variants that enable both causal supervision and robustness evaluation. To achieve adaptive reasoning, MedCausalX employs a two-stage reflection mechanism with \texttt{<causal>} and \texttt{<verify>} tokens to enable autonomous transitions from preliminary causal analysis to verification and correction. Finally, a hierarchical causal optimization pipeline sequentially integrates trajectory-level supervised learning with reinforcement learning on error-attributed trajectories, enforcing causal consistency across the reasoning process and disentangling genuine causal dependencies from shortcut associations.

%Evaluation on multiple medical imaging benchmarks demonstrates that MedCausalX significantly improves compositional reasoning capabilities, increasing localization IoU by 12.3\%, achieving diagnostic consistency of 89.7\% (15.8 percentage points higher than Med-CoT), reducing hallucination rate by 43.6\%, while outperforming state-of-the-art methods by 2-5 percentage points across three benchmarks. 
Extensive evaluation across multiple medical imaging benchmarks demonstrates that MedCausalX substantially improves compositional reasoning capabilities, enhancing diagnostic consistency, spatial localization, and hallucination suppression while consistently surpassing state-of-the-art methods. These results signify a paradigm shift from correlation-driven pattern matching toward causally grounded and trustworthy multimodal medical AI. The main contributions of this paper are as follows:

\begin{itemize}
\item 
We build CRMed, a dataset containing fine-grained causal reasoning chains and counterfactual variants, designed to support causal supervision in medical vision-language models.

\item We introduce a reflection-driven causal reasoning paradigm that enables models to \emph{detect and correct causal disruptions} during inference, transforming static chain-of-thought reasoning into an adaptive process.
\item We design a unified learning framework that \emph{aligns reasoning trajectories with causal semantics}, integrating trajectory-level supervision with error-attributed reinforcement learning to suppress shortcut associations.
\item MedCausalX achieves consistent gains across diverse medical benchmarks, substantially improving localization accuracy, diagnostic consistency, and interpretability, marking a step toward clinically reliable multimodal AI.

\end{itemize}

\section{Related Works}
\label{sec:related}

\noindent\textbf{Chain-of-Thought and Self-Reflection.} 
Chain-of-thought (CoT) enhances reasoning by decomposing complex problems into intermediate steps~\cite{wei2022chain,kojima2022large,wang2024chain,lin2025gamebot}. Some methods further improve CoT through diverse path sampling and majority voting~\cite{wang2022self} and demonstrates that CoT can emerge without explicit prompting via alternative decoding strategies~\cite{wang2024chain}. However, these methods struggle with autonomous self-correction without external feedback~\cite{huang2023large,ChenLSZ24}, occasionally degrading after intrinsic correction attempts. Training-based approaches show promise via reinforcement learning~\cite{kumar2024training} and reflection mechanisms~\cite{shinn2023reflexion,moskvoretskii2025self,NEURIPS2023_91edff07}. In medical contexts, reflection-based methods enable physician-like reasoning via hypothesis validation~\cite{huang2025medreflect}. Recent works have further improved VLM reasoning via structured 
knowledge integration and spatial 
understanding~\cite{wei2023kicgpt,wei2026dynamicgtr,xie2026spatialqa,
hou2025codev,ma2026thinkingblueprintsassistingvisionlanguage}. Yet existing methods lack mechanisms to dynamically determine when to invoke causal reasoning. Our reflection-token mechanism enables adaptive reasoning selection and explicit causal verification.
%However, current approaches lack mechanisms to dynamically determine when causal reasoning should be invoked. Our reflection-token mechanism enables adaptive reasoning selection and explicit verification of causal decomposition.

\vspace{1mm}
\noindent\textbf{Causal Reasoning in Medical Imaging.} 
Causal reasoning addresses fundamental limitations of correlation-based learning that remains vulnerable to spurious associations and dataset shift~\cite{zhao2026non,castro2020causality,geirhos2020shortcut,liu2024intrinsic}. VLMs have been widely 
applied across various domains~\cite{chen2024360+x,lu2026gastricxmultimodalmultiphasebenchmark,fu2026neurosymb,wang2025spatialclip,zhou2026unifiedthinkergeneralreasoning,song2025hume,deng2025best3dscenerepresentation,
deng2025gaussiandwm3dgaussiandriving,yao2024visual,li2025stitchfusion,li2025maris,li2025exploring,li2025taco, li2025miv, li2025catp}, and in the medical field, 
VLMs have advanced diagnostic reasoning by jointly 
processing visual and textual 
information~\cite{hu2023advancing,pan2025medvlm,li2025mmt,liu2025m,liu2024scanext, jiang2024med, jiang2025hulu}, with CoT prompting showing promise in multi-step clinical reasoning~\cite{lai2026med,liu2024medcot}. Existing studies mitigate confounding through statistical interventions~\cite{xu2025structure} or counterfactual modeling~\cite{cai2024counterfactual}, yet these methods are limited to specific question-answering setups. Our framework introduces structured causal annotations and controlled interventions that enable systematic modeling from anatomical localization through pathological characterization to diagnostic inference.

\vspace{1mm}
\noindent\textbf{Reinforcement Learning Optimization.} 
Reinforcement learning aligns large language models through iterative policy refinement~\cite{ouyang2022training,bai2022constitutional,zeng2025janusvln,zeng2025futuresightdrive,lin2026surgical,tan2025bottom,zhou2026look,zhou2025dropping}. Reinforcement Learning from Human Feedback (RLHF) establishes the foundational paradigm with supervised fine-tuning, reward modeling, and policy optimization~\cite{christiano2017deep,ziegler2019fine,schulman2017proximal,wang2025care,yang2026evotool,yangtooltree}. Efficient alternatives include Direct Preference Optimization (DPO)~\cite{rafailov2023direct}, and Group Relative Policy Optimization (GRPO), which improves efficiency 
through group-level advantage estimation~\cite{shao2024deepseekmath,guo2025deepseek,hou2025codev}. Recent analyses have revealed theoretical limitations of group-based methods,  including biased advantage estimation~\cite{yang2026your,jintic,pang2025theory}, sparse process-level guidance~\cite{ding2026prpo}, and sensitivity to  divergence choices~\cite{li2025choice}.
% and  Group Relative Policy Optimization (GRPO), which improves efficiency through group-level advantage estimation.
In medical reasoning, composite reward functions incorporating diagnostic accuracy, causal consistency, and format validity have shown promise~\cite{chen2024huatuogpt,pan2025medvlm,lai2026med,Li2025Efficient}, with 
contrast-driven approaches further improving reward 
reliability~\cite{liu2026cdrrm}. Existing systems often apply uniform corrections~\cite{liu2026automated}, whereas our error-attributed policy optimization effectively localizes causal failures and builds targeted preference pairs for fine-grained refinement.

\begin{figure*}
    \centering
    \includegraphics[width=1\linewidth]{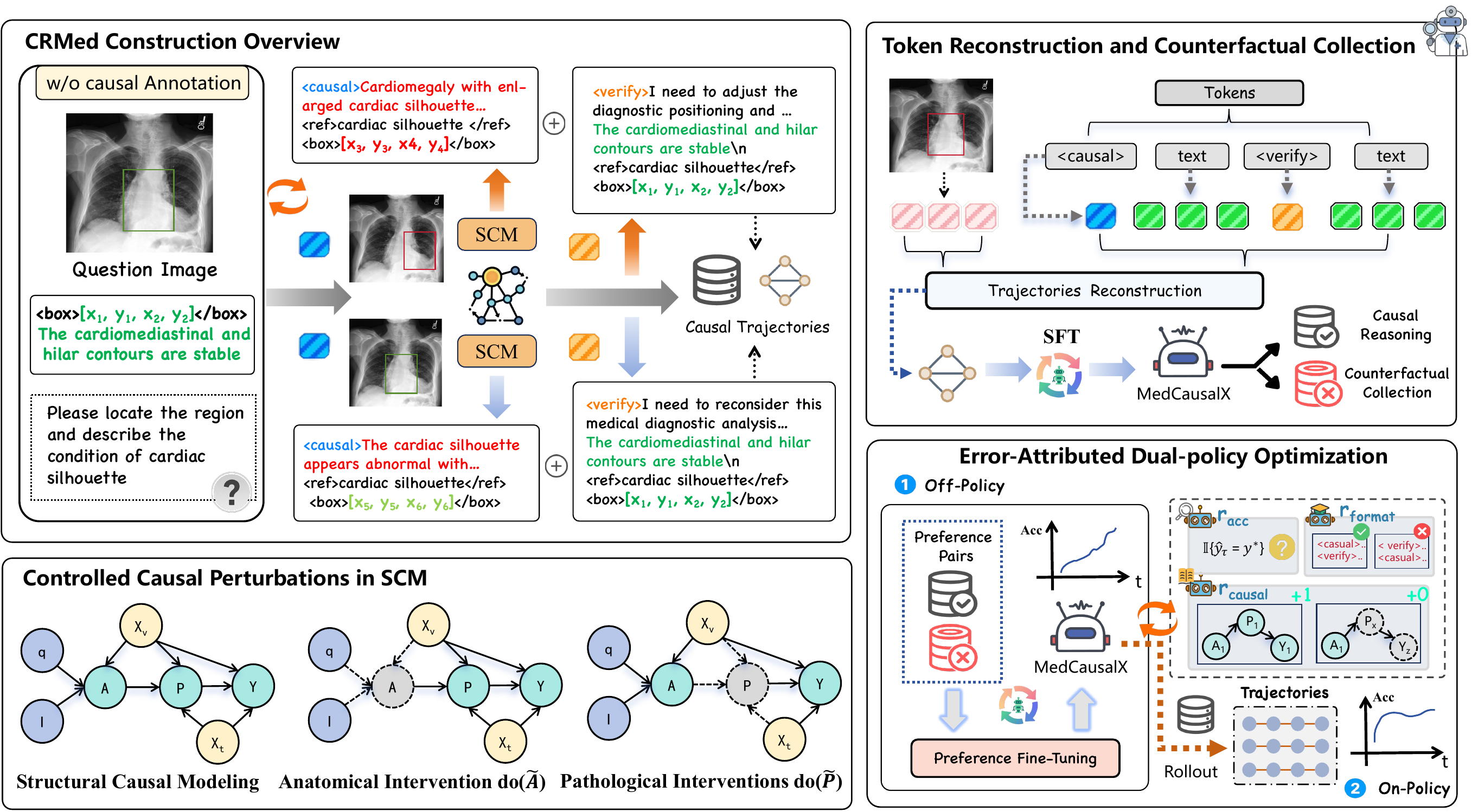}
	% \caption{Overview of the MedCausalX framework. CRMed dataset provides fine-grained causal annotations and counterfactual samples for supervision. The framework factorizes clinical reasoning via a structural causal model (SCM), employs token-level reconstruction for two-stage reflection. The reasoning trajectories are subsequently optimized via a dual-policy optimization scheme that couples off-policy preference alignment (DPO) with on-policy causal reinforcement (GRPO), bridging stability and adaptability in causal reasoning refinement.}    
    \caption{Overview of MedCausalX framework. We first construct CRMed dataset via fine-grained annotations, then factorize reasoning via a structural causal model (SCM) with reflective tokens (\texttt{<causal>}, \texttt{<verify>}) enabling two-stage adaptive reasoning. After causal Supervised Fine-Tuning (SFT), MedCausalX is subsequently optimized via dual-policy that couples off-policy preference alignment (DPO) with on-policy causal reinforcement (GRPO), bridging stability and adaptability to enhance the model's causal reasoning capability.}
	\label{fig:overall}
\end{figure*}
\section{Methodology}
\label{sec:method}

\subsection{Inductive Causal Reasoning via CRMed}
\label{sec:dataset}

We aim to address two critical limitations of current Med-CoT approaches: first, reasoning annotations are often post-hoc rationalizations generated by LLMs rather than faithful reflections of clinical reasoning, which may introduce systematic biases and reinforce spurious associations; second, existing datasets typically lack fine-grained anatomical grounding, limiting their ability to support causal analysis in medical imaging contexts.

\noindent\textbf{Structured causal factorization.}
We model diagnostic reasoning as a structured conditional generation process over three semantically grounded variables: anatomical annotation $A$, pathological characterization $P$, and diagnostic conclusion $Y$. 
% Conditioned on the medical image $I$ and clinical query $q$, we factorize the \emph{joint conditional distribution} as
Conditioned on medical image $I$ and clinical query $q$, we factorize the \emph{joint conditional distribution} as
\begin{equation}
\small
p_\theta(Y,P,A \mid I,q)
=
p_\theta(Y \mid A,P,I,q)\cdot p_\theta(P \mid A,I,q)\cdot p_\theta(A \mid I,q).
\label{eq:joint_factorization}
\end{equation}
This factorization reflects a clinically motivated reasoning order from anatomical grounding to pathological interpretation and finally diagnostic decision.

\noindent\textbf{Structural causal model.}
We formulate diagnostic reasoning under a structural causal model (SCM):
\begin{equation}
\mathcal{M} = \langle \mathcal{V}, \mathcal{U}, \mathcal{F}, P(\mathcal{U}) \rangle,
\end{equation}
where $\mathcal{V} = \{A, P, Y\}$ ,
$\mathcal{U} = \{X_v, X_t, U_c\}$ denotes the exogenous variables in the SCM, where $X_v$ and $X_t$ denote latent variability in the visual and textual domains respectively, while $U_c$ represents unobservable confounders that persist throughout the reasoning chain.

The deterministic structural functions $\mathcal{F} = {\Phi_L, \Phi_P, \Phi_D}$ define the generative mechanisms:
\begin{equation}
\begin{split}
    A &= \Phi_L(I, q, X_v, U_c), \\
    P &= \Phi_P(A, X_v, X_t, U_c), \\
    Y &= \Phi_D(P, X_v, X_t, U_c).
\end{split}
\label{eq:scm_functions}
\end{equation}
We assume the exogenous variables are mutually independent, i.e., $P(\mathcal{U}) = P(X_v) P(X_t) P(U_c)$. This SCM induces a directed acyclic graph (DAG) where the primary clinical flow $A \to P \to Y$ is subject to spurious correlations induced by the confounders.

\noindent\textbf{Approximate interventions via localized perturbations.}
To distinguish genuine causal reasoning from shortcut correlations, CRMed constructs intervention-inspired samples through \emph{localized perturbations} that selectively alter anatomical or pathological factors while keeping unrelated evidence as stable as possible.
For anatomy-targeted perturbations, we modify spatial grounding signals, \textit{e.g.}, replacing bounding box annotations or synthesizing counterfactual variants with altered anatomical structures.
For pathology-targeted perturbations, we modify pathological attributes within a fixed anatomical region, \textit{e.g.}, by replacing pathology labels under the same localization or rewriting the pathological component in the reasoning chain.
We denote the resulting perturbed input as $I^{\text{int}}$.

These perturbations are intended as \emph{proxy interventions} that approximate manipulations of the corresponding semantic variables (anatomical or pathological) under a locality assumption, rather than exact Pearlian interventions on the endogenous variables.
Accordingly, they should be interpreted as practical approximations to $do(A{=}\tilde{a})$ or $do(P{=}\tilde{p})$, rather than exact structural interventions that replace the underlying structural equations.

\noindent\textbf{Interventional sample sets.}
To separate robust causal reasoning from spurious pattern matching, CRMed incorporates three types of samples:
\begin{align}
\mathcal{D}_{\text{shortcut}} &= \{(I^{\text{int}}, q, \tilde{a}, p, C^{\text{spur}}, y)\}, \\
\mathcal{D}_{\text{partial}}  &= \{(I^{\text{int}}, q, a, \tilde{p}, C^{\text{flaw}}, y)\}, \\
\mathcal{D}_{\text{causal}}   &= \{(I, q, a, p, C^{\checkmark}, y)\},
\end{align}
where $\tilde{a}$ and $\tilde{p}$ denote perturbed anatomical and pathological factors, respectively, $C^{\text{spur}}$ indicates shortcut-prone reasoning chains, $C^{\text{flaw}}$ denotes logically inconsistent pathological analysis, and $C^{\checkmark}$ represents clinically grounded causal reasoning chains.

\noindent\textbf{Reflective token augmentation for unified reasoning.}
Clinical diagnosis relies on iterative self-examination: practitioners continuously revisit and revise hypotheses upon recognizing potential errors, rather than proceeding linearly from observation to conclusion~\cite{zhou2025explainable}. To instantiate this reflective process, we augment the base vocabulary $\mathcal{V}_0$ with reflective tokens $\mathcal{T} = \{\langle\texttt{CAUSAL}\rangle, \langle\texttt{VERIFY}\rangle\}$, enabling dynamic transitions between preliminary causal analysis and verification-based correction.

During training, all samples in CRMed follow a unified sequence structure:
\begin{equation}
\mathcal{S} = (I, q, \langle\texttt{CAUSAL}\rangle, R^{\text{biased}}, \langle\texttt{VERIFY}\rangle, R^{\text{gold}}, y),
\end{equation}
where $R^{\text{biased}} \in \{C^{\text{spur}}, C^{\text{flaw}}\}$ represents reasoning chains containing spurious correlations or logical inconsistencies from $\mathcal{D}_{\text{shortcut}}$ or $\mathcal{D}_{\text{partial}}$, and $R^{\text{gold}} = C^{\checkmark}$ denotes clinically validated ground-truth causal chains from $\mathcal{D}_{\text{causal}}$. This construction enforces a two-stage reasoning protocol: the model first generates a preliminary causal analysis conditioned on $\langle\texttt{CAUSAL}\rangle$, then performs verification and correction triggered by $\langle\texttt{VERIFY}\rangle$, ultimately producing refined reasoning chains with enhanced causal consistency. This format enables the model to learn adaptive causal correction through contrastive exposure to both erroneous and correct reasoning trajectories within a unified training paradigm.

\subsection{Causal Reasoning Optimization}
%\section{Training the Causal Reasoning}
We adopt a two-stage optimization that explicitly attributes causal errors and strengthens reasoning robustness:

%After Stage I establishes initial causal reasoning via supervised fine-tuning on CRMed, Stage II focuses on eliminating remaining spurious correlations through progressive optimization.

\subsubsection{Preliminary: SFT with Causal Decomposition}
\label{sec:stage1}

In this stage, we establish foundational causal reasoning and self-correction  capabilities through supervised fine-tuning on the CRMed dataset, maximizing the joint likelihood of generating anatomical localization, causal reasoning chains, and diagnostic predictions, conditioned on the medical image-query pair.
%\begin{equation}
%\label{eq:sft}
%\mathcal{L}_{\text{SFT}}(\theta) = -\mathbb{E}_{(I,q,R,y)\sim\mathcal{D}_{\text{CRMed}}}[\log p_\theta(R, y | I, q)],
%\end{equation}
\begin{equation}
\small
\label{eq:sft_trajectory}
\mathcal{L}_{\text{SFT}}(\theta) = -\mathbb{E}_{(I,q,L,C,y)\sim\mathcal{D}_{\text{}}}\left[\log p_\theta(L, C, y \mid I, q, \mathcal{T})\right],
\end{equation}
where $\mathcal{T}$ denotes the reflective token conditioning the reasoning mode, $L$ represents the anatomical localization, $C$ the causal chain, and $y$ the final diagnostic label.

\subsubsection{Refinement: Error-Attributed Optimization}\label{sec:stage2}
%We further employ reinforcement learning to refine causal reasoning after supervised fine-tuning (Eq.~\ref{eq:sft_trajectory}).
%Specifically, we decouple the optimization into two complementary stages:
%(i) \emph{off-policy preference optimization} using Direct Preference Optimization (DPO)~\cite{rafailov2023direct} to exploit previously collected erroneous trajectories, enabling stable policy refinement from human-like preference signals without costly online sampling; and
%(ii) \emph{on-policy reinforcement learning} via Group Relative Policy Optimization (GRPO)~\cite{shao2024deepseekmath} to further align the model with causal-consistent reasoning through online exploration and self-improvement.
%This hybrid scheme stabilizes learning by leveraging DPO’s sample efficiency for error correction while retaining GRPO’s capacity to adaptively improve reasoning policies under true rollouts, achieving fine-grained causal robustness. We empirically observe that initializing GRPO with DPO-aligned policy significantly stabilizes on-policy updates and accelerates convergence (see Supplementary).
After supervised trajectory fine-tuning (Eq.~\ref{eq:sft_trajectory}), we further refine causal reasoning via reinforcement learning.
The training is divided into two complementary stages: an off-policy preference optimization stage for efficient error correction and an on-policy reinforcement learning stage for adaptive causal refinement.
%Details of both training paradigms are provided below.

\paragraph{Stage I: Off-policy Training.}
We leverage collected error trajectories 
$\mathcal{D}_{\text{error}} = \{(x^i_{I\&T}, Y^i_{\text{err}}, Y^i_{\text{corr}})\}_{i=1}^{N_e}$ 
to construct preference pairs for Direct Preference Optimization (DPO). 
For each input $x_{I\&T}$, we enforce $Y_{\text{corr}} \succ Y_{\text{err}}$ and optimize:
\begin{equation}
\small
\scalebox{0.9}{$
\begin{aligned}
\mathcal{L}_{\text{DPO}}(\pi_\theta; \pi_{\text{ref}}) 
&= -\mathbb{E}_{(x,Y_w,Y_l) \sim \mathcal{D}_{\text{error}}} \Big[ 
    \log \sigma \Big( 
        \beta \log \frac{\pi_\theta(Y_w | x)}{\pi_{\text{ref}}(Y_w | x)} \\
&\quad - \beta \log \frac{\pi_\theta(Y_l | x)}{\pi_{\text{ref}}(Y_l | x)} 
    \Big) 
\Big],
\end{aligned}$}
\end{equation}
where $Y_w = Y_{\text{corr}}$ denotes winning trajectory, $Y_l = Y_{\text{err}}$, $\pi_\theta$ is the trainable policy, $\pi_{\text{ref}}$ the reference policy, $\beta > 0$ the KL coefficient, and $\sigma(\cdot)$ the sigmoid function.

\noindent\textbf{Error Localization Mechanism.}
To enhance preference pair informativeness, we precisely identify the failure point $t_{\text{fail}}$ via step-wise semantic divergence detection:
\begin{equation}
% t_{\text{fail}} = \min \left\{ t \in [1, T] \,:\, \mathcal{S}(Y^t_{\text{err}}, Y^t_{\text{gt}}) < \tau \right\},
t_{\text{fail}} = \min \left\{ t \in \{1, 2\} \,:\, \mathcal{S}(Y^t_{\text{err}}, Y^t_{\text{gt}}) < \tau \right\},
\label{eq:error_localization}
\end{equation}
where $\mathcal{S}(\cdot, \cdot): \mathcal{Y} \times \mathcal{Y} \to [0,1]$ measures semantic similarity and $\tau \in (0,1)$ is a divergence threshold. 
We then contrast the continuations $Y^{\text{err}}_{\geq t_{\text{fail}}}$ and $Y^{\text{corr}}_{\geq t_{\text{fail}}}$ while conditioning on the shared prefix $Y^{\text{corr}}_{< t_{\text{fail}}}$, concentrating learning on critical divergence points and improving sample efficiency.

\paragraph{Stage II: On-policy Training.}
Building on error-localized and high-quality samples, we adopt Group Relative Policy Optimization (GRPO) for on-policy refinement. 
For each input $(x, y^*)$, we sample $G$ trajectories $\{\tau_g\}_{g=1}^G$ from $\pi_\theta$ and compute a composite reward $R^{(g)} = R_{\text{total}}(\tau_g, y^*)$ measuring correctness and reasoning quality:
\begin{equation}
\small
\mathcal{L}_{\text{GRPO}} = -\mathbb{E}_{\{\tau_g\}_{g=1}^G \sim \pi_{\theta}} \left[\frac{1}{G} \sum_{g=1}^G A^{(g)} \log \pi_{\theta}(\tau_g \mid x)\right],
\label{eq:grpo_loss}
\end{equation}
with group-relative advantage $A^{(g)} = (R^{(g)} - \bar{R}) / (\sigma_R + \epsilon)$, 
where $\bar{R}$ and $\sigma_R$ are the mean and standard deviation of rewards within the group. 
This intra-group normalization stabilizes updates and reduces variance, improving efficiency.

\begin{table*}[t]
\centering
\Huge
\renewcommand{\arraystretch}{1.1}
\setlength{\tabcolsep}{2.8pt}
\caption{Comprehensive quantitative comparison on diverse region-centric tasks, including region-to-text identification and text-to-region detection. Abbreviations in this Table: Obj-F1 denotes Object F1, Reg-F1 denotes Region F1, Align-F1 denotes Alignment F1, and Acc denotes Accuracy. The best results are highlighted in \textbf{bold}, while the second-best are \underline{underlined}. }
\label{tab:region_centric_comprehensive}
\resizebox{1\textwidth}{!}{%
\begin{tabular}{l|ccccc|ccccc|cccc|cccc}
\toprule[1.2pt]\midrule
\multirow{3}{*}{\textbf{Method}} & \multicolumn{10}{c|}{\textbf{Region-to-Text Identification}} & \multicolumn{8}{c}{\textbf{Text-to-Region Detection}} \\
\cmidrule(lr){2-11} \cmidrule(lr){12-19}
& \multicolumn{5}{c|}{\textbf{Structure Identification}} & \multicolumn{5}{c|}{\textbf{Lesion Identification}} & \multicolumn{4}{c|}{\textbf{Single-Region Detection}} & \multicolumn{4}{c}{\textbf{Multi-Region Detection}} \\
\cmidrule(lr){2-6} \cmidrule(lr){7-11} \cmidrule(lr){12-15} \cmidrule(lr){16-19}
& \textbf{BLEU-1} & \textbf{F1} & \textbf{Recall} & \textbf{BertScore} & \textbf{Acc} & \textbf{BLEU-1} & \textbf{F1} & \textbf{Recall} & \textbf{BertScore} & \textbf{Acc} & \textbf{Obj-F1} & \textbf{Reg-F1} & \textbf{Align-F1} & \textbf{IoU} & \textbf{Obj-F1} & \textbf{Reg-F1} & \textbf{Align-F1} & \textbf{IoU} \\
\midrule
Qwen2.5-VL~\cite{bai2025qwen25vl} & 0.15 & 0.25 & 0.58 & 52.34 & 0.12 & 0.14 & 0.23 & 0.55 & 48.56 & 0.11 & 57.82 & 8.15 & 6.23 & 17.45 & 49.35 & 5.87 & 4.65 & 14.82 \\
InternVL-2~\cite{chen2024expanding} & 0.13 & 0.21 & 0.52 & 51.78 & 0.11 & 0.13 & 0.20 & 0.51 & 47.92 & 0.10 & 56.60 & 7.24 & 5.45 & 16.61 & 48.23 & 5.41 & 4.18 & 13.99 \\
Med-Flamingo~\cite{moor2023med} & 0.13 & 0.72 & 4.90 & 24.87 & 0.01 & 0.12 & 0.68 & 4.72 & 23.45 & 0.01 & 18.47 & 2.35 & 1.82 & 5.23 & 12.68 & 1.54 & 1.18 & 3.47 \\
LLaVA-Med~\cite{li2023llava} & 0.43 & 1.15 & 10.69 & 34.53 & 1.74 & 0.38 & 1.05 & 9.85 & 32.18 & 1.52 & 24.83 & 3.76 & 2.94 & 7.85 & 18.95 & 2.42 & 1.87 & 5.36 \\
RadFM~\cite{wu2023towards} & 0.35 & 0.80 & 4.75 & 37.07 & 1.88 & 0.32 & 0.75 & 4.42 & 35.62 & 1.68 & 21.56 & 3.12 & 2.38 & 6.74 & 15.82 & 1.98 & 1.52 & 4.58 \\
MedDr~\cite{he2024meddr} & 0.75 & 1.27 & 3.52 & 50.28 & 1.36 & 0.68 & 1.18 & 3.28 & 48.56 & 1.24 & 28.34 & 4.25 & 3.47 & 9.16 & 22.47 & 2.85 & 2.24 & 6.42 \\
MedRegA~\cite{wang2024interpretable} & \underline{78.34} & \underline{78.95} & \underline{79.02} & \textbf{91.63} & \underline{73.06} & \underline{61.09} & \underline{61.90} & \textit{63.07} & \textbf{82.62} & \underline{59.42} & \underline{77.93} & \underline{45.11} & \underline{36.53} & \underline{42.70} & \underline{68.52} & \underline{24.78} & \underline{22.36} & \underline{31.63} \\
\midrule\midrule
%\rowcolor[RGB]{231, 248, 254}MedCausalX-DPO & \underline{79.82} & \underline{80.15} & \underline{81.25} & 88.45 & \underline{75.15} & \underline{63.27} & \underline{64.35} & \underline{65.80} & 81.90 & \underline{61.35} & \underline{82.45} & \underline{46.82} & \underline{44.92} & \underline{44.35} & \underline{72.38} & \underline{29.51} & \underline{26.82} & \underline{35.76} \\
\rowcolor[RGB]{231, 248, 254}\textbf{MedCausalX} & \textbf{80.11} & \textbf{80.47} & \textbf{81.63} & \underline{88.72} & \textbf{75.52} & \textbf{63.65} & \textbf{64.78} & \textbf{66.02} & \underline{82.10} & \textbf{61.72} & \textbf{82.83} & \textbf{47.05} & \textbf{45.12} & \textbf{44.68} & \textbf{72.75} & \textbf{29.83} & \textbf{27.01} & \textbf{36.02} \\
\midrule
\bottomrule[1.2pt]
\end{tabular}%
}
\end{table*}

\begin{table*}[t]
\centering
\renewcommand{\arraystretch}{1.15}
\setlength{\tabcolsep}{5.2pt}
\caption{Performance comparison on medical VQA benchmarks with causal reasoning evaluation. Metrics include answer Accuracy (Acc), Diagnostic Consistency (Diag-C), and Hallucination Rate (Hall, lower is better). Results are reported as mean$_{\text{std}}$ over 5-fold cross-validation. Best results in \textbf{bold}, second-best \underline{underlined}. '*' denotes fine-tuned models.}
\label{tab:medical_vqa_results}
\resizebox{1\textwidth}{!}{%
\begin{tabular}{l|ccc|ccc|ccc|ccc|ccc}
\toprule[1.2pt]\hline
\multirow{2}{*}{\textbf{Method}} & \multicolumn{3}{c|}{\textbf{SLAKE}~\cite{liu2021slake}} & \multicolumn{3}{c|}{\textbf{VQA-RAD}~\cite{lau2018dataset}} & \multicolumn{3}{c|}{\textbf{PathVQA}~\cite{he2020pathological}} & \multicolumn{3}{c|}{\textbf{PMC-VQA}~\cite{zhang2023pmc}} & \multicolumn{3}{c}{\textbf{Average}} \\
& \textbf{Acc} & \textbf{Diag-C} & \textbf{Hall$\downarrow$} & \textbf{Acc} & \textbf{Diag-C} & \textbf{Hall$\downarrow$} & \textbf{Acc} & \textbf{Diag-C} & \textbf{Hall$\downarrow$} & \textbf{Acc} & \textbf{Diag-C} & \textbf{Hall$\downarrow$} & \textbf{Acc} & \textbf{Diag-C} & \textbf{Hall$\downarrow$} \\
\hline
\rowcolor[RGB]{240,240,240}\multicolumn{16}{c}{\textbf{General Vision-Language Models}} \\
\hline
Qwen2.5-VL~\cite{bai2025qwen25vl} & 68.5$_{1.8}$ & 58.3$_{2.1}$ & 62.5$_{1.9}$ & 52.4$_{2.3}$ & 54.2$_{2.0}$ & 65.8$_{1.7}$ & 48.3$_{2.1}$ & 52.8$_{1.9}$ & 68.2$_{2.2}$ & 58.2$_{1.6}$ & 56.7$_{1.8}$ & 63.5$_{2.0}$ & 56.9$_{1.9}$ & 55.5$_{1.9}$ & 65.0$_{1.9}$ \\
InternVL-2~\cite{chen2024expanding} & 70.8$_{1.6}$ & 61.2$_{1.8}$ & 59.8$_{2.1}$ & 55.1$_{2.0}$ & 57.5$_{1.9}$ & 62.5$_{1.8}$ & 52.7$_{1.9}$ & 55.3$_{2.0}$ & 65.3$_{2.1}$ & 61.5$_{1.5}$ & 59.2$_{1.7}$ & 60.8$_{1.9}$ & 60.0$_{1.8}$ & 58.3$_{1.9}$ & 62.1$_{2.0}$ \\
\hline
\rowcolor[RGB]{240,240,240}\multicolumn{16}{c}{\textbf{Medical-Specific Vision-Language Models}} \\
\hline
Med-Flamingo~\cite{moor2023med} & 33.2$_{2.7}$ & 38.5$_{2.5}$ & 75.2$_{2.3}$ & 44.6$_{2.6}$ & 42.8$_{2.4}$ & 71.8$_{2.5}$ & 47.9$_{2.4}$ & 45.6$_{2.3}$ & 68.5$_{2.6}$ & 35.9$_{2.8}$ & 40.5$_{2.5}$ & 73.8$_{2.4}$ & 40.4$_{2.6}$ & 41.9$_{2.4}$ & 72.3$_{2.5}$ \\
LLaVA-Med~\cite{li2023llava} & 45.6$_{2.3}$ & 46.8$_{2.2}$ & 68.5$_{2.4}$ & 27.4$_{2.8}$ & 40.2$_{2.5}$ & 74.5$_{2.3}$ & 58.2$_{2.0}$ & 53.2$_{2.1}$ & 63.8$_{2.2}$ & 63.8$_{1.8}$ & 55.5$_{2.0}$ & 65.2$_{2.1}$ & 48.8$_{2.2}$ & 48.9$_{2.2}$ & 68.0$_{2.3}$ \\
RadFM~\cite{wu2023towards} & 82.3$_{1.2}$ & 70.8$_{1.4}$ & 51.2$_{1.8}$ & 60.6$_{1.7}$ & 63.5$_{1.5}$ & 55.8$_{1.9}$ & 40.2$_{2.4}$ & 50.8$_{2.1}$ & 66.5$_{1.8}$ & 45.8$_{2.2}$ & 56.2$_{1.9}$ & 62.8$_{1.7}$ & 57.2$_{1.9}$ & 60.3$_{1.7}$ & 59.1$_{1.8}$ \\
MedDr~\cite{he2024meddr} & 83.4$_{1.1}$ & 72.5$_{1.3}$ & 48.5$_{1.7}$ & \underline{78.5}$_{1.3}$ & 70.2$_{1.4}$ & 50.2$_{1.8}$ & 62.8$_{1.6}$ & 64.8$_{1.5}$ & 56.8$_{1.9}$ & 66.2$_{1.5}$ & 66.5$_{1.4}$ & 54.5$_{1.8}$ & 72.7$_{1.4}$ & 68.5$_{1.4}$ & 52.5$_{1.8}$ \\
MedRegA~\cite{wang2024interpretable} & \underline{84.1}$_{1.0}$ & \underline{73.8}$_{1.2}$ & \underline{47.8}$_{1.6}$ & 76.9$_{1.2}$ & \underline{69.5}$_{1.3}$ & \underline{50.8}$_{1.7}$ & \underline{68.5}$_{1.4}$ & \underline{68.5}$_{1.3}$ & \underline{52.5}$_{1.8}$ & \underline{79.5}$_{1.1}$ & \underline{72.8}$_{1.2}$ & \underline{49.8}$_{1.6}$ & \underline{77.3}$_{1.2}$ & \underline{71.2}$_{1.3}$ & \underline{50.2}$_{1.7}$ \\
\hline
\rowcolor[RGB]{240,240,240}\multicolumn{16}{c}{\textbf{Medical Chain-of-Thought Models}} \\
\hline
MedCoT~\cite{liu2024medcot}* & 83.9$_{1.1}$ & 71.5$_{1.3}$ & 49.5$_{1.7}$ & 75.8$_{1.3}$ & 68.2$_{1.4}$ & 52.2$_{1.8}$ & 66.2$_{1.5}$ & 66.8$_{1.4}$ & 54.8$_{1.9}$ & 80.3$_{1.0}$ & 71.8$_{1.2}$ & 50.5$_{1.7}$ & 76.6$_{1.2}$ & 69.6$_{1.3}$ & 51.8$_{1.8}$ \\
MedVLM-R1~\cite{pan2025medvlm}* & 85.7$_{0.9}$ & 75.5$_{1.1}$ & 45.2$_{1.5}$ & 77.3$_{1.2}$ & 71.8$_{1.2}$ & 47.5$_{1.6}$ & 70.8$_{1.3}$ & 70.5$_{1.2}$ & 49.8$_{1.7}$ & 82.4$_{0.9}$ & 75.2$_{1.1}$ & 45.8$_{1.5}$ & 79.1$_{1.1}$ & 73.3$_{1.2}$ & 47.1$_{1.6}$ \\
Med-R1~\cite{lai2026med}* & 85.4$_{0.9}$ & 74.8$_{1.1}$ & 46.5$_{1.6}$ & 76.9$_{1.2}$ & 70.5$_{1.3}$ & 48.8$_{1.7}$ & 69.5$_{1.4}$ & 69.2$_{1.3}$ & 51.5$_{1.8}$ & 81.9$_{1.0}$ & 74.5$_{1.1}$ & 47.2$_{1.6}$ & 78.4$_{1.1}$ & 72.3$_{1.2}$ & 48.5$_{1.7}$ \\
\hline\hline
\rowcolor[RGB]{231, 248, 254}\textbf{MedCausalX (Ours)} & \textbf{87.2}$_{\textbf{0.8}}$ & \textbf{81.5}$_{\textbf{1.0}}$ & \textbf{34.2}$_{\textbf{1.4}}$ & \textbf{79.8}$_{\textbf{1.1}}$ & \textbf{77.3}$_{\textbf{1.1}}$ & \textbf{36.5}$_{\textbf{1.5}}$ & \textbf{73.2}$_{\textbf{1.2}}$ & \textbf{75.8}$_{\textbf{1.1}}$ & \textbf{38.9}$_{\textbf{1.6}}$ & \textbf{84.6}$_{\textbf{0.8}}$ & \textbf{80.2}$_{\textbf{1.0}}$ & \textbf{35.8}$_{\textbf{1.4}}$ & \textbf{81.2}$_{\textbf{1.0}}$ & \textbf{78.7}$_{\textbf{1.1}}$ & \textbf{36.4}$_{\textbf{1.5}}$ \\
\hline
\bottomrule[1.2pt]
\end{tabular}%
}
\end{table*}

\noindent\textbf{Composite Reward Function.} 
The overall reward is formulated as
$
R(\tau) = R_{\text{acc}}(\tau, y^*) + R_{\text{format}}(\tau) + R_{\text{causal}}(\tau),
$
integrating three complementary components that jointly evaluate reasoning quality from predictive accuracy, structural validity, and causal coherence.

\textit{(1) Accuracy Reward: } 
$
R_{\text{acc}}(\tau, y^*) = \mathbb{I}\{\hat{y}_{\tau} = y^*\},
$
where $\hat{y}_{\tau}$ and $y^*$ denote the model’s output and the correct label, respectively. This term assesses whether the final prediction matches the ground truth.

\textit{(2) Format Reward}  is used to ensure structural integrity:
\begin{equation}
R_{\text{format}}(\tau) = \frac{1}{|\tau|} \sum_{t=1}^{|\tau|} \mathbb{I}\{w_t \in \mathcal{V}_{\text{valid}}\},
\end{equation}
where $|\tau|$ is the trajectory length, $w_t$ the $t$-th token, and $\mathcal{V}_{\text{valid}}$ the predefined valid token set. We assess whether correction or verification tokens follow the desired format.

\textit{(3) Causal Consistency Reward} evaluates the logical coherence across consecutive reasoning steps:
\begin{equation}
R_{\text{causal}}(\tau) = \frac{1}{|\mathcal{S}_{\tau}| - 1} 
\sum_{i=1}^{|\mathcal{S}_{\tau}| - 1} \text{Cons}(s_i, s_{i+1}),
\end{equation}
where $\mathcal{S}_{\tau} = \{s_1, \ldots, s_{|\mathcal{S}_{\tau}|}\}$ represents the reasoning sequence extracted from trajectory $\tau$, 
and $\text{Cons}(s_i, s_{i+1})$ quantifies the causal consistency between consecutive steps.

Together, these rewards encourage predictions that are accurate, structurally well-formed, and causally coherent, ensuring that model refinements not only correct errors but also preserve the integrity of underlying causal reasoning.

\section{Experiments}

\subsection{Experimental Setup}

\vspace{1mm}
\noindent
\textbf{Baselines.}
We compare our method with three comprehensive categories of baselines:
(1) General Vision-Language Models, including InternVL-2~\cite{chen2024expanding}, GPT-4V~\cite{hurst2024gpt}, and Qwen2.5-VL~\cite{bai2025qwen25vl}, which exhibit strong multimodal understanding capabilities;
(2) Medical-Specific VLMs, such as Med-Flamingo~\cite{moor2023med}, LLaVA-Med~\cite{li2023llava}, RadFM~\cite{wu2023towards}, MedDr~\cite{he2024meddr}, BiomedGPT~\cite{zhang2024generalist}, and MedRegA~\cite{wang2024interpretable}, representing the state-of-the-art in medical vision-language understanding; and
(3) Medical Chain-of-Thought Models, including MedVLM-R1~\cite{pan2025medvlm} and Med-R1~\cite{lai2026med}, which incorporate CoT reasoning processes for interpretability.

\vspace{1mm}
\noindent
\textbf{Comparison Benchmarks.}
We evaluate MedCausalX across five medical imaging benchmarks. \textit{MIMIC-CXR}~\cite{johnson2019mimic} contains chest X-ray image-report pairs for report generation. \textit{SLAKE}~\cite{liu2021slake} and \textit{VQA-RAD}~\cite{lau2018dataset} provide multi-modality radiology images with clinical question-answer pairs for visual question answering. \textit{PathVQA}~\cite{he2020pathological} focuses on histopathological image reasoning. \textit{SA-Med2D-20M}~\cite{ye2023sa} supports two region-centric tasks with pixel-level annotations: {Region-to-Text Identification} requires recognizing anatomical structures given bounding boxes, while {Text-to-Region Detection} requires localizing specific organs or anomalies from textual descriptions. All experiments use 5-fold cross-validation for statistical reliability.

% \vspace{1mm}
% \noindent
% \textbf{CRMed Dataset.} 
% We construct CRMed by augmenting existing benchmarks with three annotation layers. First, fine-grained anatomical localization provides bounding boxes for diagnostically relevant structures. Second, structured causal chains decompose each diagnosis into anatomical observation ($A$), pathological characterization ($P$), and diagnostic conclusion ($Y$) following our SCM formulation. Third, counterfactual variants are generated via controlled interventions: $\mathcal{D}_{\text{shortcut}}$ contains incorrect anatomical localizations to test spatial robustness, while $\mathcal{D}_{\text{partial}}$ includes logically inconsistent pathological characterizations. This design enables both supervised causal training and systematic robustness evaluation under interventional distributions.

\begin{figure*}
    \centering
    \includegraphics[width=1\linewidth]{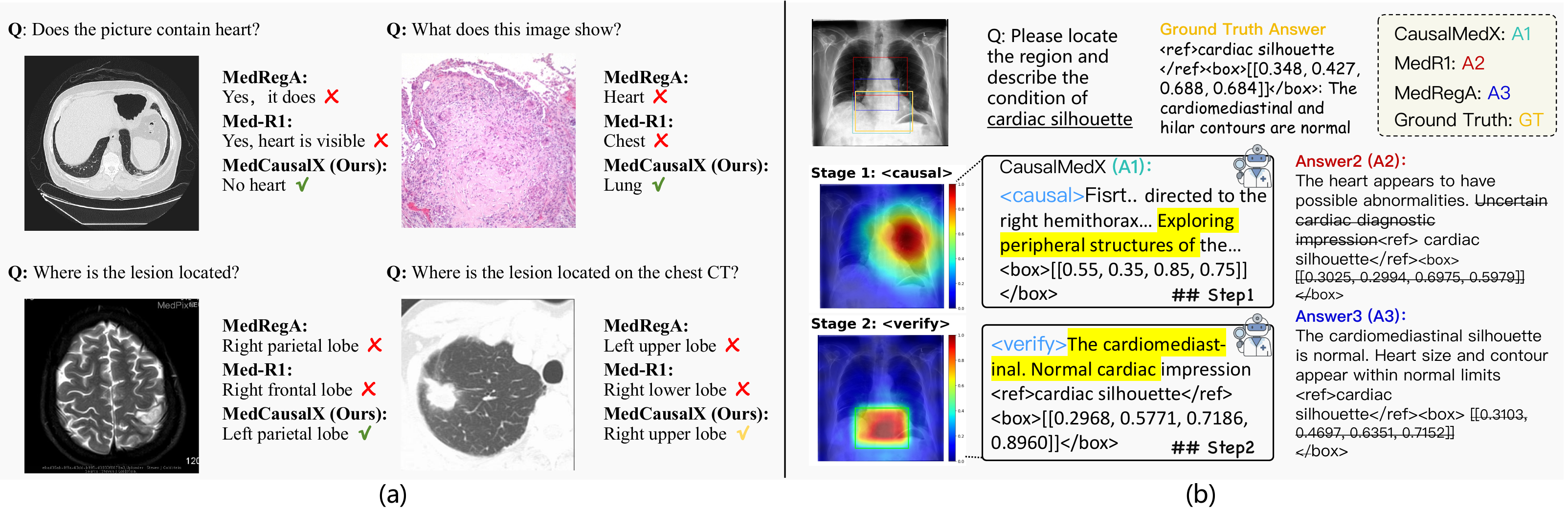}
	\caption{Comparison of causal reasoning in medical VQA and region-grounded diagnosis. (a) MedRegA and Med-R1 exhibit spurious correlations and mistakes, while MedCausalX captures correct predictions via structured causal decomposition. (b) Two-stage adaptive causal reasoning and verification, producing refined bounding boxes and logically consistent diagnostic descriptions. Attention maps are used to visualize the model's focus. GT: ground truth; A1--A3: predictions from MedCausalX, Med-R1, and MedRegA, respectively.}
	\label{fig:qualitative_cases}
\end{figure*}

\vspace{1mm}
\noindent \textbf{Implementation Details and Evaluation Metrics.} MedCausalX builds upon Qwen2.5-VL architecture~\cite{bai2025qwen25vl} with model variants ranging from 3B to 32B parameters. Our main experiments utilize the 32B variant for optimal causal reasoning capabilities. Training follows a two-stage paradigm: Stage I performs supervised fine-tuning on CRMed utilizing causal decomposition objectives (Eq.~\ref{eq:sft_trajectory}) for 3 epochs; Stage II applies preference optimization, combining off-policy DPO (2 epochs) with on-policy GRPO (2000 steps) to enhance causal consistency. Error localization employs semantic similarity threshold $\tau=0.7$ for failure point identification. We use AdamW optimizer~\cite{loshchilov2017decoupled,pan2023toward,jin2026adam} ($lr=1\times10^{-6}$) and train on 6 NVIDIA A100 GPUs with mixed precision. The causal consistency function $\text{Cons}(s_i, s_{i+1})$ is implemented via an LLM-as-judge protocol: we prompt GPT-4o to assess whether consecutive reasoning steps maintain valid causal dependencies and whether the corrective action triggered by $\langle\texttt{VERIFY}\rangle$ successfully rectifies the identified causal disruption, returning a binary score that is averaged across step pairs. The CRMed dataset augments existing benchmarks with three annotation layers: fine-grained anatomical localization (bounding boxes), structured causal chains (SCM), and reconstructed variants ($\mathcal{D}_{\text{shortcut}}$, $\mathcal{D}_{\text{partial}}$).

The Evaluation metrics spanning three categories: (1) \emph{Linguistic and answer quality}: BLEU-1, F1-score, Recall, BertScore and Accuracy (Acc) assess textual fluency and answer correctness; (2) \emph{Spatial accuracy}: IoU, Region Accuracy, Object-F1, and Region-F1 quantify localization precision; (3) \emph{Multimodal alignment}: Align-F1 and Align Accuracy measure visual-textual correspondence. Moreover, \emph{Hallucination Rate} calculates the fraction of generated entities absent from image and ground-truth.

\subsection{Main Results}

\subsubsection{Region-Centric Spatial Reasoning}
Table~\ref{tab:region_centric_comprehensive} evaluates bidirectional region-text alignment. MedCausalX achieves best performance on 16 of 18 metrics across structure/lesion identification and single-/multi-region detection. Notably, we attain 29.83\% Region-F1 on multi-region detection, outperforming both general and medical VLMs. This stems from causal decomposition that enforces anatomical localization prior to pathological characterization, rather than relying on shortcut correlations. While MedRegA maintains text fluency advantages, our method surpasses it in spatial grounding (F1: +8.59\%), revealing that previous models prioritize linguistic plausibility whereas our causal supervision enforces faithful spatial-semantic correspondence crucial in multi-lesion scenarios.

\subsubsection{Medical VQA with Causal Reasoning}
Table~\ref{tab:medical_vqa_results} demonstrates MedCausalX's consistent superiority across four medical VQA benchmarks. Compared to MedVLM-R1, our method achieves notable improvements in accuracy (81.2\% vs 79.1\%) and reduces hallucination rate (47.1\%→36.4\%). Trajectory-level optimization prevents contradictions between intermediate steps and final predictions. Even fine-tuned CoT models like MedCoT exhibit high hallucination, confirming that conventional sequential reasoning cannot suppress spurious correlations. Training with CRMed effectively discriminates genuine causal dependencies from dataset artifacts, while structured causal factorization grounds predictions in precise anatomical evidence rather than global image statistics.

\begin{table}[t]
\centering
\caption{Performance comparison on chest X-ray report generation benchmark. Our method employs two-stage causal reasoning with adaptive verification based on MedCausalX-32B model. Best results are highlighted in \textbf{bold}, second-best are \underline{underlined}.}
\label{tab:mimic_cxr_results}
\resizebox{0.475\textwidth}{!}{
\begin{tabular}{l|c|c|c|>{\centering\arraybackslash}p{1.35cm}}
\toprule[1pt]\hline
\textbf{Method} & \textbf{BLEU-1} & \textbf{Region Acc} & \textbf{Align Acc} & \textbf{IoU}\\
\midrule[0.6pt]
Qwen2.5-VL~\cite{bai2025qwen25vl} & 18.43 & 22.30 & 8.15 & 17.45\\
InternVL~\cite{chen2024internvl} & 19.46 & 57.00 & 12.00 & 38.00\\
Med-Flamingo~\cite{moor2023med} & 21.67 & - & - & - \\
MedDr~\cite{he2024meddr} & 31.27 & - & - & - \\
MedRegA~\cite{wang2024interpretable} & \underline{33.18} & \underline{76.59} & \underline{62.29} & \underline{52.07}\\
\midrule
%\rowcolor[RGB]{231, 248, 254}MedCausalX-DPO & 35.62 & 78.30 & 63.20 & 52.09\\
\rowcolor[RGB]{231, 248, 254}\textbf{MedCausalX}& \textbf{37.18} & \textbf{81.12} & \textbf{67.38} & \textbf{55.71}\\
\hline
\bottomrule[1pt]
\end{tabular}
}
\end{table}

\subsubsection{Grounded Report Generation}
\label{mimic}
Table~\ref{tab:mimic_cxr_results} evaluates MIMIC-CXR report generation. We achieve 37.18\% BLEU-1, substantially outperforming MedRegA across linguistic, spatial, and alignment metrics. Causal factorization successfully bridges visual evidence and textual generation, while trajectory-level optimization ensures anatomically grounded descriptions rather than linguistically fluent but spatially inconsistent reports. This demonstrates that MedCausalX generalizes to X-ray report generation with precise spatial grounding, highlighting the transferability of its causal reasoning framework.

\begin{table}[t]
\centering
\caption{Systematic ablation of core components and training stage variants on MIMIC-CXR benchmark evaluates each module's contribution to causal reasoning and verification capabilities.}
\label{tab:comprehensive_ablation}
\resizebox{0.475\textwidth}{!}{
\begin{tabular}{>{\raggedright\arraybackslash}p{3.4cm}|c|c|c|>{\centering\arraybackslash}p{1.35cm}}
\toprule[1pt] \hline
\textbf{Configuration}
& \textbf{BLEU-1} & \textbf{Region Acc} & \textbf{Align Acc} & \textbf{IoU}\\
\midrule[0.6pt]
\rowcolor[RGB]{231, 248, 254}\textbf{MedCausalX (Full)} & \textbf{37.18} & \textbf{81.12} & \textbf{67.38} & \textbf{55.71}\\
\midrule[0.6pt]
\multicolumn{5}{l}{\textit{Main Component Ablation}} \\
\xmark~CRMed Dataset & 26.47 & 41.20 & 18.35 & 26.55 \\
\xmark~Reflective Tokens & 29.93 & 51.20 & 32.50 & 38.46\\
\xmark~Causal SFT & 26.02 & 43.80 & 21.75 & 28.55\\
\xmark~RL Training & 26.96 & 63.50 & 47.30 & 38.09\\
\xmark~Error Collection & 32.02 & 75.20 & 58.45 & 49.55\\
\midrule[0.7pt]
\multicolumn{5}{l}{\textit{Training Stage Variants}} \\
Base Model & 23.15 & 35.00 & 12.00 & 24.49\\
+ Causal SFT & 32.45 & 58.30 & 41.25 & 38.62\\
+ DPO & 35.62 & 78.30 & 63.20 & 52.09\\
+ GRPO & \textbf{37.18} & \textbf{81.12} & \textbf{67.38} & \textbf{55.71} \\
\hline
\bottomrule[1pt] 
\end{tabular}
}
\end{table}

\subsection{Indepth Case Analysis}
Figure~\ref{fig:qualitative_cases} demonstrates MedCausalX's superior causal reasoning through contrastive examples. Panel (a) reveals baseline failures: MedRegA misidentifies anatomical structures via global pattern matching, while Med-R1 produces spatially inconsistent localizations. MedCausalX correctly rejects spurious associations through structured $A \rightarrow P \rightarrow Y$ decomposition. Panel (b) visualizes our reflection mechanism with attention heatmaps and reasoning trajectories. The $\langle$causal$\rangle$ token initiates preliminary spatial analysis on relevant anatomical regions, and subsequent $\langle$verify$\rangle$ invocation refines localization with causally coherent descriptions. This dynamic correction, absent in static chain-of-thought baselines, enables adaptive error detection and grounds reasoning in genuine anatomical evidence rather than superficial image patterns.

\begin{figure}
    \centering
    \includegraphics[width=1\linewidth]{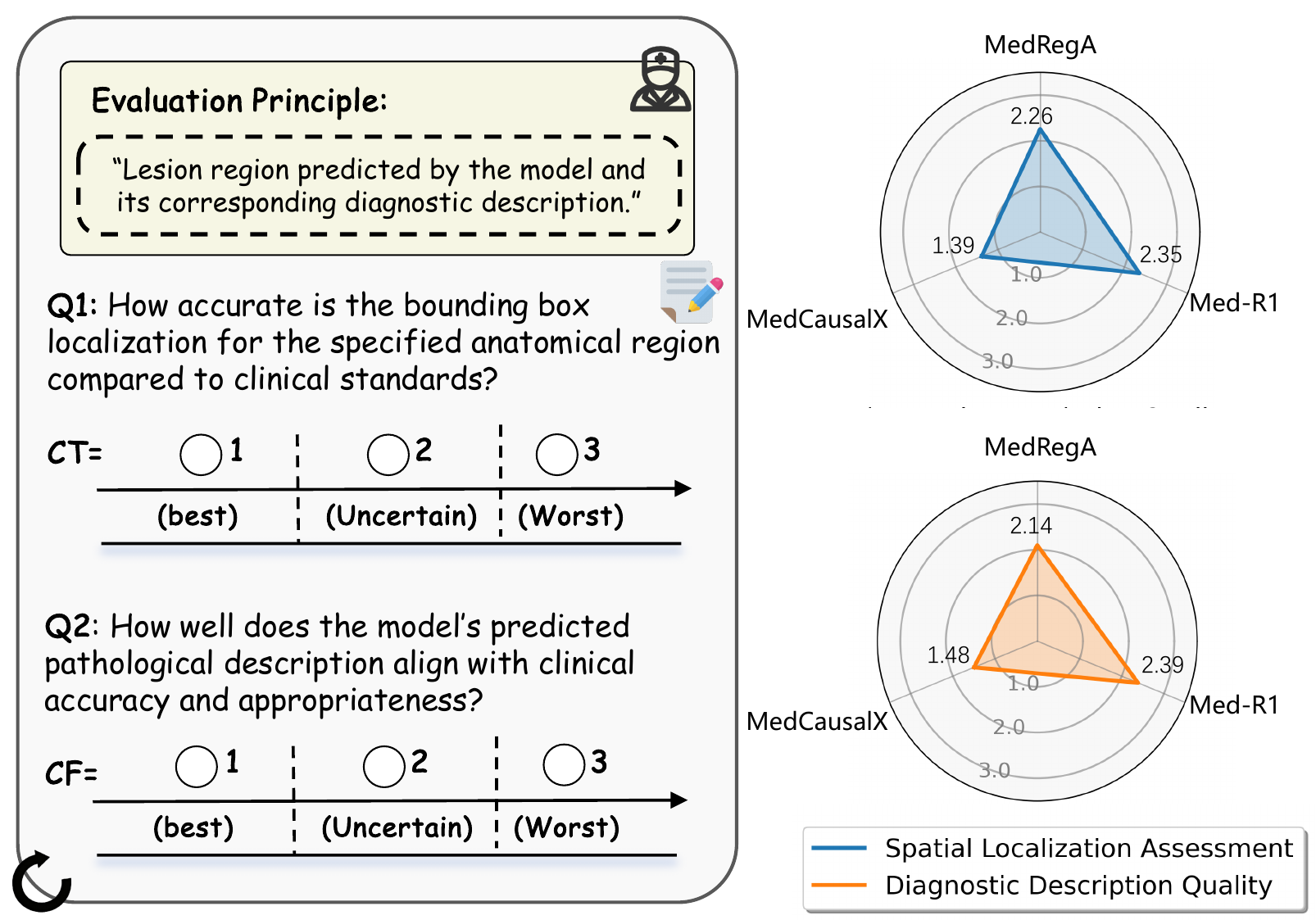}
    \caption{Radiologists' evaluation and comparative results. Left: 3-point clinical criteria for spatial localization (Q1), diagnostic description (Q2). Right: radar chart comparing MedRegA, Med-R1, and MedCausalX on localization (blue) and description (orange); lower scores indicate better performance.}
    \label{fig:exp2}
\end{figure}
% Analysis section - single paragraph

\subsection{Medical Expert Evaluation}
We conducted an expert evaluation with board-certified radiologists with over 20 years of experience, assessing model outputs for X-ray report grounding (Section~\ref{mimic}) on standardized 3-point scales for spatial localization and diagnostic quality. The evaluation protocol (left panel) was designed to reflect clinical practice, with stringent criteria for anatomical precision and pathological characterization. Radar chart analysis (right panel) reveals MedCausalX’s superior performance, with mean scores of 1.39 for spatial localization and 1.48 for diagnostic quality. The results provide strong expert endorsement, confirming that explicit causal factorization yields outputs that are both anatomically precise and diagnostically coherent, supporting the reliability of our causal reasoning framework.

\begin{table}[t]
\centering
\caption{Hyperparameter sensitivity analysis. Selected configuration (shaded) achieves optimal balance across metrics.}
\label{tab:hyperparameter_ablation}
\resizebox{0.475\textwidth}{!}{
\begin{tabular}{>{\raggedright\arraybackslash}p{2.6cm}|>{\centering\arraybackslash}p{1.4cm}|c|c|>{\centering\arraybackslash}p{1.35cm}}
\toprule[1pt]
\hline
\textbf{Param.}
& \textbf{BLEU-1} & \textbf{Region Acc} & \textbf{Align Acc} & \textbf{IoU}\\
\midrule[0.6pt]
\multicolumn{5}{l}{\textit{Error Localization Threshold ($\tau$)}} \\
$\tau = 0.5$ & 33.47 & 77.85 & 63.91 & 52.74\\
$\tau = 0.6$ & 35.12 & 79.67 & 66.24 & 54.83\\
\rowcolor[RGB]{231, 248, 254}$\tau = 0.7$ (default) & \textbf{37.18} & \textbf{81.12} & \textbf{67.38} & \textbf{55.71}\\
$\tau = 0.8$ & 33.23 & 80.17 & 65.82 & 54.96\\
$\tau = 0.9$ & 34.91 & 80.34 & 61.15 & 52.68\\
\midrule[0.7pt]
\multicolumn{5}{l}{\textit{GRPO Group Size ($G$)}} \\
$G = 4$ & 33.84 & 78.92 & 64.67 & 53.29\\
\rowcolor[RGB]{231, 248, 254}$G = 8$ (default) & \textbf{37.18} & \textbf{81.12} & 67.38 & \textbf{55.71}\\
$G = 16$ & \text{36.75} & 80.53 & \textbf{68.19} & 55.14\\
$G = 32$ & 34.91 & 79.28 & 65.43 & 53.85\\
\hline
\bottomrule[1pt]
\end{tabular}
}
\end{table}

\subsection{Ablation Studies}

\paragraph{Component Contributions.}
Table~\ref{tab:comprehensive_ablation} quantifies each module's impact. Removing CRMed causes severe degradation (BLEU-1: 37.18→26.47), confirming structured causal supervision is indispensable for learning genuine $A!\rightarrow!P!\rightarrow!Y$ dependencies. Excluding reflective tokens reduces performance to 29.93\%, as static pipelines cannot adaptively invoke verification. Ablating RL training shows supervised learning alone cannot distinguish causally valid from spurious chains. Progressive stages reveal complementary roles: Causal SFT establishes foundations, while two-stage RL exploits collected errors for stable off-policy learning and further performs adaptive exploration.

\vspace{1mm} \noindent \textbf{Hyperparameter Robustness Analysis.} Table~\ref{tab:hyperparameter_ablation} examines key parameter sensitivity. The error localization threshold $\tau$ follows an inverted-U pattern. Overly sensitive detection with $\tau=0.5$ introduces false positives that corrupt preference pairs, whereas conservative thresholds with $\tau=0.9$ miss subtle errors. The optimal value of $\tau=0.7$ balances precision and recall. For the GRPO group size $G$, a value of 8 balances variance reduction and computational cost. Smaller groups tend to underestimate advantages, while larger groups lead to diminishing returns. Stability across $G \in [8,16]$ demonstrates practical robustness and indicates that minimal tuning is required for deployment.

\section{Conclusion}

%This paper proposes MedCausalX, a unified framework that explicitly models and optimizes causal reasoning within medical vision-language inference. Through the structured causal factorization ($A\!\rightarrow\!P\!\rightarrow\!Y$) with reflective token architecture and error-attributed dual-policy optimization combining off-policy DPO with on-policy GRPO, MedCausalX realizes the transition from correlation-driven shortcuts to causally verifiable reasoning chains. Experimental results show that MedCausalX's adaptive reflection mechanism enables dynamic error detection and correction during inference, significantly outperforming existing medical CoT methods across multiple benchmarks in localization accuracy, diagnostic consistency, and hallucination reduction. The counterfactual robustness analysis reveals that interventional training with $\mathcal{D}_{\text{shortcut}}$ and $\mathcal{D}_{\text{partial}}$ effectively discriminates genuine causal dependencies from spurious correlations, demonstrating that structured causal supervision is essential for trustworthy medical AI. Our approach provides a new technical pathway for developing causally grounded reasoning-enabled medical multimodal systems.
This paper presents MedCausalX, a framework that explicitly models causal reasoning in medical vision-language inference. By combining causal factorization ($A\!\rightarrow\!P\!\rightarrow\!Y$), a reflective token architecture, and dual-policy optimization combining off-policy with on-policy RL, MedCausalX shifts from correlation-driven inference to verifiable reasoning chains. Experiments show its adaptive reflection mechanism enables dynamic error detection and correction, outperforming existing medical CoT methods in localization, diagnostic consistency, and hallucination reduction. Training with reconstructed $\mathcal{D}_{\text{shortcut}}$ and $\mathcal{D}_{\text{partial}}$ effectively distinguishes genuine causal dependencies from spurious correlations, highlighting the necessity of causal supervision in trustworthy clinical multimodal systems.

\section{Additional Implementation Details}

\subsection{Detailed Experimental Configuration}

MedCausalX employs Qwen2.5-VL-32B~\cite{bai2025qwen25vl} as the base architecture with parameter-efficient LoRA adaptation targeting query, key, and value projection layers. Data augmentation preserves diagnostic integrity through random rotation ($\pm 15°$), horizontal flipping (probability 0.5), brightness/contrast adjustment ($\pm 20\%$), and Gaussian noise injection ($\sigma=0.03$). The framework augments vocabulary with reflective tokens $\langle\texttt{CAUSAL}\rangle$ and $\langle\texttt{VERIFY}\rangle$, enabling all training samples to follow unified structure $(I, q, \langle\texttt{CAUSAL}\rangle, R^{\text{biased}}, \langle\texttt{VERIFY}\rangle, R^{\text{gold}}, y)$ for learning adaptive verification through reasoning trajectories.

Training proceeds through two-stage optimization following preliminary supervised fine-tuning. Stage I (Causal SFT) maximizes joint likelihood over anatomical localization, causal chains, and diagnostic predictions for 3 epochs using AdamW with learning rate $\eta=1\times10^{-6}$, weight decay $5\times10^{-5}$, gradient clipping threshold 1.0, and 500-step linear warmup followed by cosine annealing. Stage II integrates dual-policy refinement: off-policy DPO~\cite{rafailov2023direct} over 2 epochs with KL coefficient $\beta=0.1$ leverages error trajectories where semantic similarity threshold $\tau=0.7$ identifies failure points $t_{\text{fail}}$ via $\mathcal{S}(Y^t_{\text{err}}, Y^t_{\text{gt}})$ comparison, constructing preference pairs contrasting erroneous continuations against correct trajectories; on-policy GRPO~\cite{shao2024deepseekmath} over 2000 steps samples $G=8$ trajectories per input, computing group-relative advantages $A^{(g)} = (R^{(g)} - \bar{R}) / (\sigma_R + \epsilon)$ with stability constant $\epsilon=1\times10^{-8}$. Distributed training across 6 NVIDIA A100 (40GB) GPUs with gradient accumulation (4 steps) and mixed precision (bfloat16) converges within approximately 56 hours per dataset using 5-fold cross-validation with stratified 8:1:1 splits. Table~\ref{tab:hyperparameters_medcausalx} details the complete configuration.

\begin{figure}[t]
    \centering
    \includegraphics[width=\linewidth]{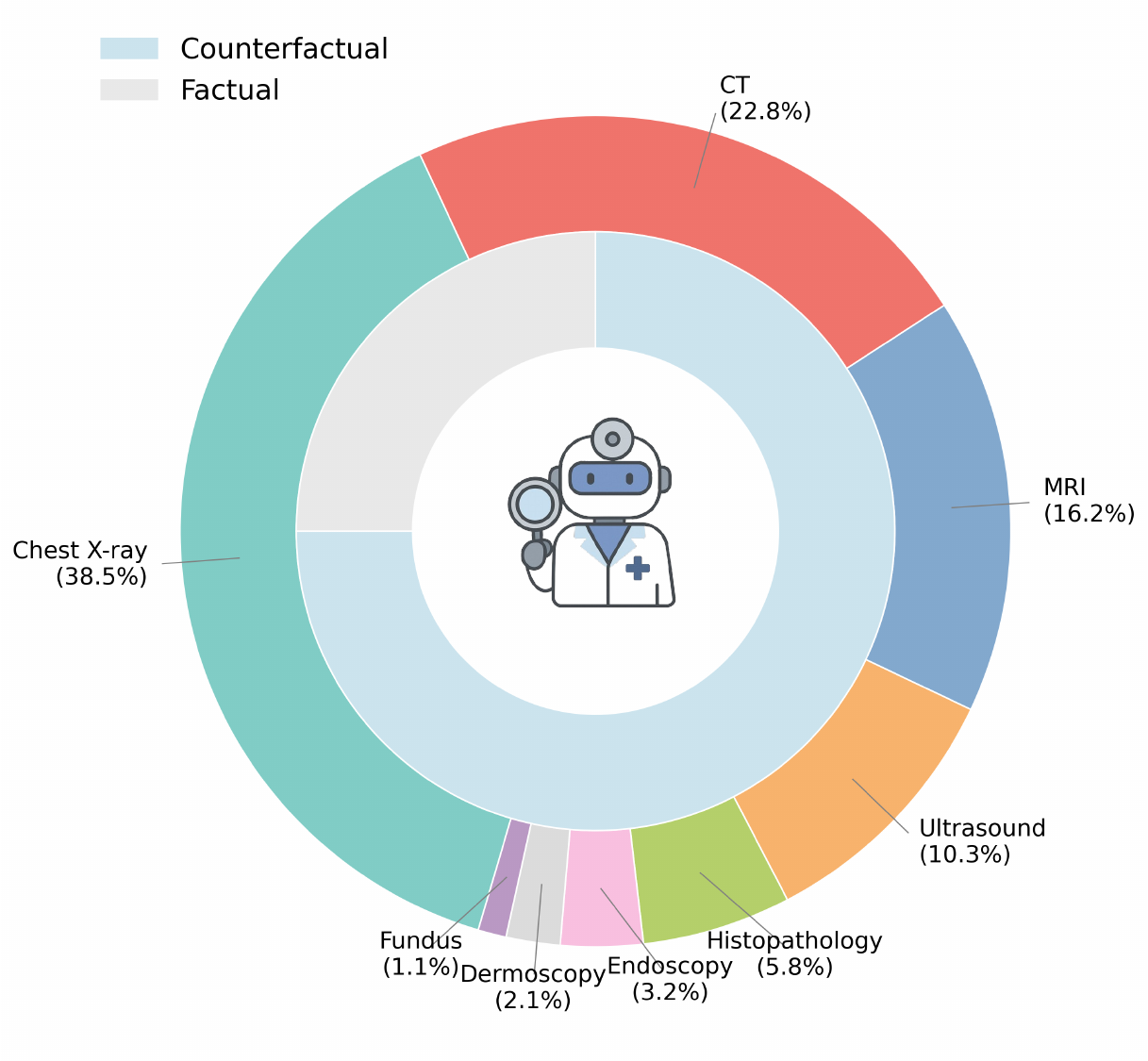}
\caption{Modality distribution and factual-counterfactual composition of CRMed dataset. The dataset spans multiple imaging modalities with constructed counterfactual variants through controlled interventions, enabling robust causal reasoning evaluation.}
    \label{fig:efficiency}
\end{figure}

\begin{table}[h]
\centering
\small
\caption{Complete hyperparameter configuration for MedCausalX training and inference.}
\label{tab:hyperparameters_medcausalx}
\begin{tabular}{l|l}
\toprule
\hline
\textbf{Hyperparameter} & \textbf{Value} \\
\midrule
Base Model & Qwen2.5-VL-32B \\
LoRA Rank ($r$) & 64 \\
LoRA Alpha ($\alpha$) & 128 \\
LoRA Dropout & 0.05 \\
Target Modules & q\_proj, k\_proj, v\_proj \\
Reflective Tokens & $\langle\texttt{CAUSAL}\rangle$, $\langle\texttt{VERIFY}\rangle$ \\
Nucleus Sampling ($p$) & 0.9 \\
Temperature ($T$) & 0.7 \\
Batch Size & 64 \\
Max Tokens ($T_{\max}$) & 2048 \\
\midrule
\multicolumn{2}{l}{\textit{Causal SFT (Stage I)}} \\
Epochs & 3 \\
Optimizer & AdamW \\
Learning Rate ($\eta$) & $1\times10^{-6}$ \\
Weight Decay & $5\times10^{-5}$ \\
Gradient Clipping & 1.0 \\
Warmup Steps & 500 \\
LR Schedule & Linear + Cosine \\
\midrule
\multicolumn{2}{l}{\textit{Dual-Policy Optimization (Stage II)}} \\
\textbf{Off-policy training:} & \\
\quad Epochs & 2 \\
\quad KL Coefficient ($\beta$) & 0.1 \\
\quad Error Threshold ($\tau$) & 0.7 \\
\textbf{On-policy training:} & \\
\quad Training Steps & 2000 \\
\quad Group Size ($G$) & 8 \\
\quad Stability Constant ($\epsilon$) & $1\times10^{-8}$ \\
\midrule
\multicolumn{2}{l}{\textit{Infrastructure}} \\
GPUs & $6\times$ A100 (40GB) \\
Gradient Accumulation & 4 steps \\
Mixed Precision & bfloat16 \\
Training Time & $\sim$56 hours/dataset \\
Cross-validation & 5-fold \\
Data Split & 8:1:1 \\
\hline
\bottomrule
\end{tabular}
\end{table}

\begin{table*}[t]
\centering
\renewcommand{\arraystretch}{1.15}
\setlength{\tabcolsep}{4.5pt}
\caption{Model scaling analysis across region-centric medical reasoning tasks. We evaluate MedCausalX variants with 3B, 14B, and 32B parameters to assess the impact of model capacity on causal reasoning performance. Best results are highlighted in \textbf{bold}, second-best are \underline{underlined}. All models trained with identical data and optimization strategies.}
\label{tab:model_scaling}
\resizebox{1\textwidth}{!}{%
\begin{tabular}{l|ccccc|ccccc|cccc|cccc}
\toprule[1.2pt]\hline
\multirow{3}{*}{\textbf{Model Scale}} & \multicolumn{10}{c|}{\textbf{Region-to-Text Identification}} & \multicolumn{8}{c}{\textbf{Text-to-Region Detection}} \\
\cmidrule(lr){2-11} \cmidrule(lr){12-19}
& \multicolumn{5}{c|}{\textbf{Structure Identification}} & \multicolumn{5}{c|}{\textbf{Lesion Identification}} & \multicolumn{4}{c|}{\textbf{Single-Region Detection}} & \multicolumn{4}{c}{\textbf{Multi-Region Detection}} \\
\cmidrule(lr){2-6} \cmidrule(lr){7-11} \cmidrule(lr){12-15} \cmidrule(lr){16-19}
& \textbf{BLEU-1} & \textbf{F1} & \textbf{Recall} & \textbf{BertScore} & \textbf{Acc} & \textbf{BLEU-1} & \textbf{F1} & \textbf{Recall} & \textbf{BertScore} & \textbf{Acc} & \textbf{Obj-F1} & \textbf{Reg-F1} & \textbf{Align-F1} & \textbf{IoU} & \textbf{Obj-F1} & \textbf{Reg-F1} & \textbf{Align-F1} & \textbf{IoU} \\
\hline
MedCausalX-3B & 53.42 & 54.68 & 55.20 & 52.15 & 48.20 & 38.29 & 39.12 & 40.35 & 48.65 & 35.12 & 68.45 & 21.47 & 18.23 & 24.49 & 55.18 & 14.23 & 12.47 & 19.82 \\
MedCausalX-14B & \underline{73.49} & \underline{74.23} & \underline{75.80} & \underline{54.20} & \underline{55.30} & \underline{56.43} & \underline{57.80} & \underline{59.25} & \underline{70.83} & \underline{52.80} & \underline{79.82} & \underline{34.62} & \underline{28.15} & \underline{33.75} & \underline{65.75} & \underline{19.56} & \underline{17.85} & \underline{26.18} \\
\rowcolor[RGB]{231, 248, 254}\textbf{MedCausalX-32B} & \textbf{79.82} & \textbf{80.15} & \textbf{81.25} & \textbf{88.45} & \textbf{70.15} & \textbf{63.27} & \textbf{64.35} & \textbf{65.80} & \textbf{81.90} & \textbf{61.35} & \textbf{82.45} & \textbf{46.82} & \textbf{44.92} & \textbf{44.35} & \textbf{72.38} & \textbf{29.51} & \textbf{26.82} & \textbf{35.76} \\
%\hline
%\textbf{Improvement} & \multicolumn{18}{c}{\textbf{3B→32B: \quad +26.40 BLEU-1,\quad +25.35 Reg-F1,\quad +21.95 Acc\quad (avg. across tasks)}} \\
%\hline
%\bottomrule[1.2pt]

\hline
\bottomrule
%\multicolumn{5}{l}{\small \rule{0pt}{2.5ex} \small \textbf{3B→32B: +26.40 BLEU-1,\quad +25.35 Reg-F1,\quad +21.95 Acc\quad }.} \\
\end{tabular}%
}
\end{table*}

\subsection{Benchmark Dataset Statistics}

We evaluate MedCausalX on established medical imaging benchmarks spanning multiple modalities and clinical domains. \textit{MIMIC-CXR}~\cite{johnson2019mimic} provides 377,110 chest radiographs from 227,835 imaging studies with free-text reports and 14 pathology labels for report generation tasks. \textit{SLAKE}~\cite{liu2021slake} contains 642 multi-modal images with 14,028 question-answer pairs across anatomy, modality, and pathology categories. \textit{VQA-RAD}~\cite{lau2018dataset} includes 315 radiology images with 3,515 QA pairs emphasizing clinical reasoning. \textit{PathVQA}~\cite{he2020pathological} comprises 32,799 pathology images with 6,719 QA pairs for histopathological interpretation. \textit{SA-Med2D-20M}~\cite{ye2023sa} offers 4.6 million 2D medical images with 19.7 million segmentation masks across 10 modalities and 31 organs, enabling region-centric tasks including structure identification and lesion localization with bounding box annotations. All experiments employ 5-fold cross-validation using stratified 8:1:1 train-validation-test splits for statistical reliability.

\begin{table}[t]
\centering
\large
\renewcommand{\arraystretch}{1.08}
\caption{Model scale and architecture comparison across medical VQA benchmarks. Light blue background indicates selected configuration.}
\label{tab:model_scale_ablation}
\resizebox{\columnwidth}{!}{
\begin{tabular}{l|cccc|c}
\toprule[1pt]
\hline
\textbf{Model} & \textbf{VQA-RAD} & \textbf{SLAKE} & \textbf{PathVQA} & \textbf{PMC-VQA} & \textbf{Average} \\
\midrule[0.6pt]
\multicolumn{6}{l}{\textit{Vision-Language Models (Zero-shot)}} \\
\midrule[0.6pt]
LLaVA-1.5-13B~\cite{liu2024improved} & 53.2 & 58.5 & 52.8 & 56.3 & 55.2 \\
InternVL-2-40B~\cite{chen2024expanding} & 59.8 & 65.2 & 61.5 & 67.1 & 63.4 \\
%\midrule[0.6pt]
%\multicolumn{6}{l}{\textit{Medical-Specific Models (Fine-tuned)}} \\
%\midrule[0.6pt]
MedRegA-40B~\cite{wang2024interpretable} & 76.9 & 84.1 & 68.5 & 79.5 & 77.3 \\
\midrule[0.6pt]
\multicolumn{6}{l}{\textit{Medical CoT Methods (Qwen2.5-VL-32B Backbone)}} \\
\midrule[0.6pt]
Med-R1~\cite{lai2026med} & 76.9 & 85.4 & 69.5 & 81.9 & 78.4 \\
MedVLM-R1~\cite{pan2025medvlm} & 77.3 & 85.7 & 70.8 & 82.4 & 79.1 \\
\midrule[0.6pt]
\multicolumn{6}{l}{\textit{MedCausalX Scale Variants (Qwen2.5-VL Backbone)}} \\
\midrule[0.6pt]
\rowcolor[RGB]{245, 245, 245}MedCausalX-3B & 67.8 & 75.9 & 62.5 & 68.2 & 68.6 \\
\rowcolor[RGB]{245, 245, 245}MedCausalX-7B & 73.5 & 81.8 & 67.3 & 74.9 & 74.4 \\
\rowcolor[RGB]{231, 248, 254}\textbf{MedCausalX-32B} & \textbf{79.8} & \textbf{87.2} & \textbf{73.2} & \textbf{84.6} & \textbf{81.2} \\
\hline
\bottomrule[1pt]
\end{tabular}
}
\end{table}

\noindent\textbf{CRMed Dataset Construction.}
The CRMed dataset augments existing benchmarks with two systematic annotation layers.  \textit{First}, we select images from SA-Med2D-20M and MIMIC-CXR based on multi-criteria quality assessment: (i) sufficient image resolution ($\geq$512$\times$512 pixels) and contrast quality (SNR$>$15dB), (ii) presence of diagnostically relevant anatomical structures confirmed through clinical relevance scoring ($\geq$4/5 by three board-certified radiologists). Selected images include bounding box annotations in format $(x_{\min}, y_{\min}, x_{\max}, y_{\max})$ covering target anatomical structures. \textit{Second}, counterfactual variants generate $\mathcal{D}_{\text{shortcut}}$ via spatial perturbations (random shifts $\pm50$ pixels, scale jittering $\times[0.8, 1.2]$) disrupting $A\!\rightarrow\!P$ dependencies, and $\mathcal{D}_{\text{partial}}$ through pathology misalignment breaking $P\!\rightarrow\!Y$ causal flow while preserving image content. Intervention quality is validated by ensuring IoU$>$0.7 between perturbed and original bounding boxes to maintain anatomy. 
The resulting corpus contains 89,342 images with 267,128 causal samples, averaging 4.2 reasoning steps per 
chain (SD=1.3). Modality distribution spans chest X-ray (38.5\%), CT (22.8\%), 
MRI (16.2\%), ultrasound (10.3\%), histopathology (5.8\%), endoscopy (3.2\%), 
dermoscopy (2.1\%), and fundus (1.1\%), with counterfactual plausibility 
scores $>$0.85 via GPT-4V evaluation.

%(1:3 factual-to-counterfactual ratio)

\begin{figure}[t]
    \centering
    \includegraphics[width=\linewidth]{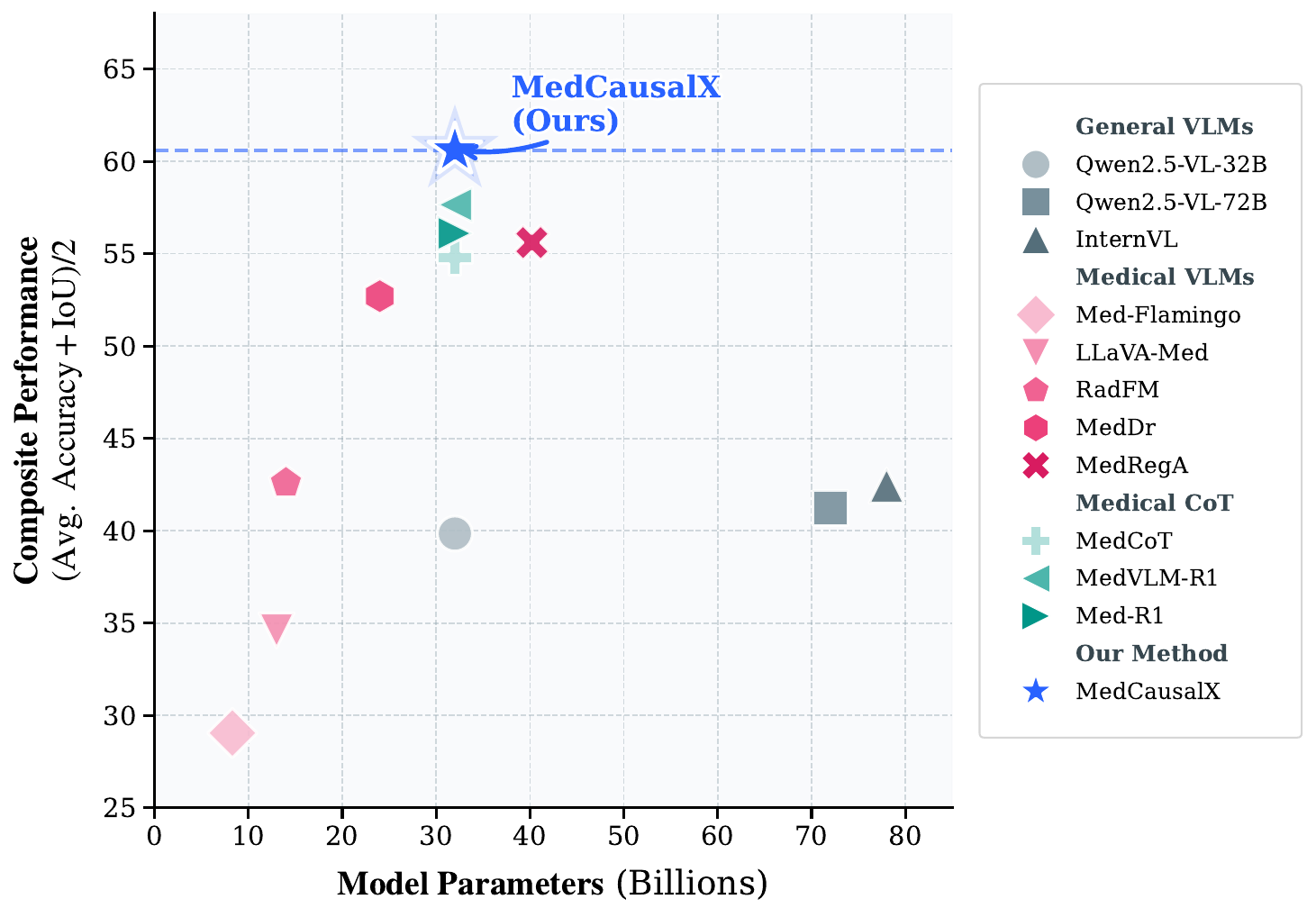}
    \caption{Comparison of model efficiency and performance across medical VLMs. MedCausalX achieves the highest composite score at 32B parameters, demonstrating superior causal reasoning capabilities while maintaining computational efficiency.}
    \label{fig:efficiency}
\end{figure}

\section{Additional Experimental Results}

\subsection{Causal Reasoning Cases}

\begin{figure*}[t]
    \centering
    \includegraphics[width=0.8\textwidth]{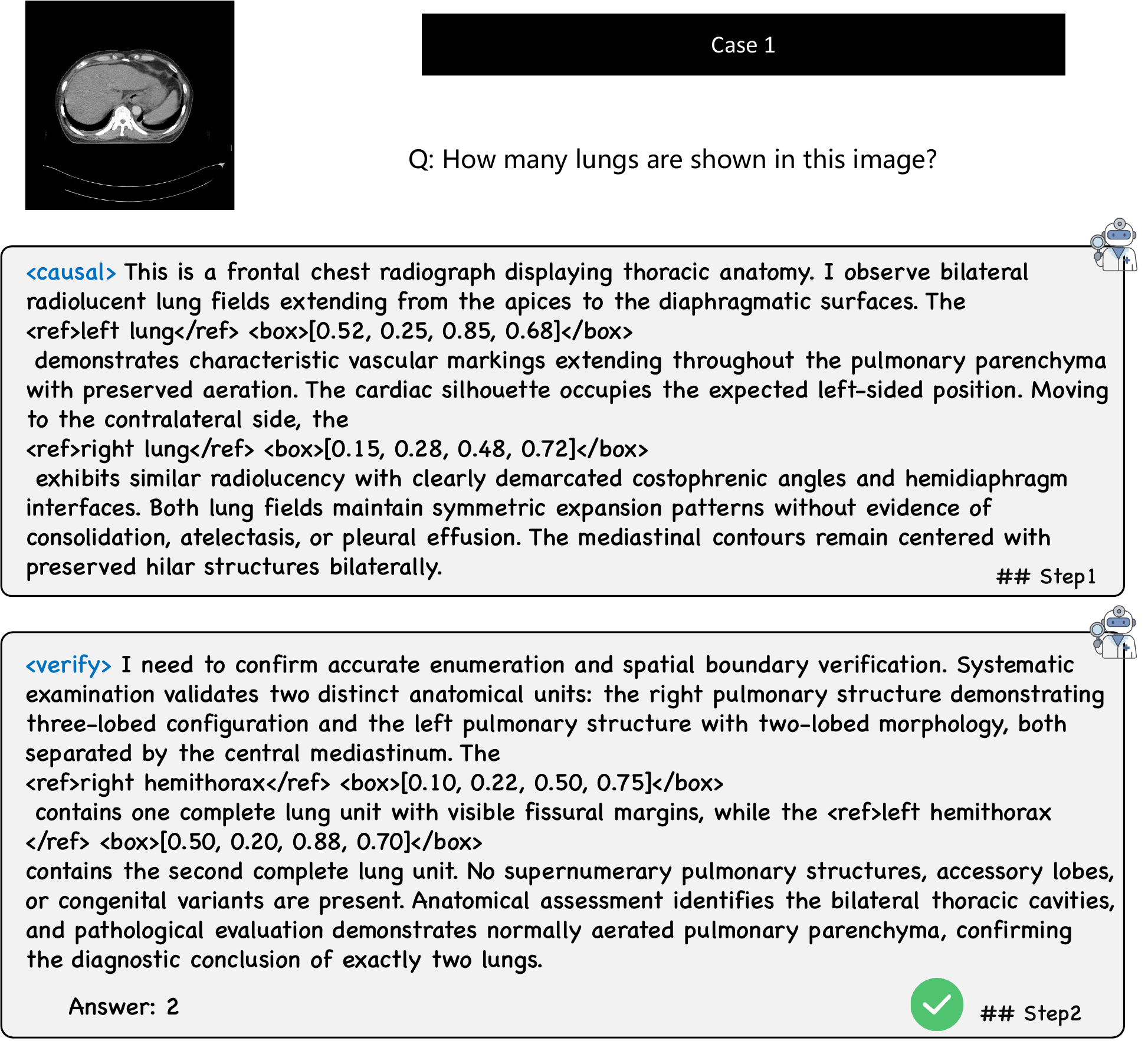}
    \caption{Chest radiograph demonstrating bilateral lung detection where MedCausalX accurately quantifies two distinct pulmonary structures through structured anatomical enumeration and spatial boundary verification.}
    \label{fig:case1_lung}
\end{figure*}

\begin{figure*}[t]
    \centering
    \includegraphics[width=0.8\textwidth]{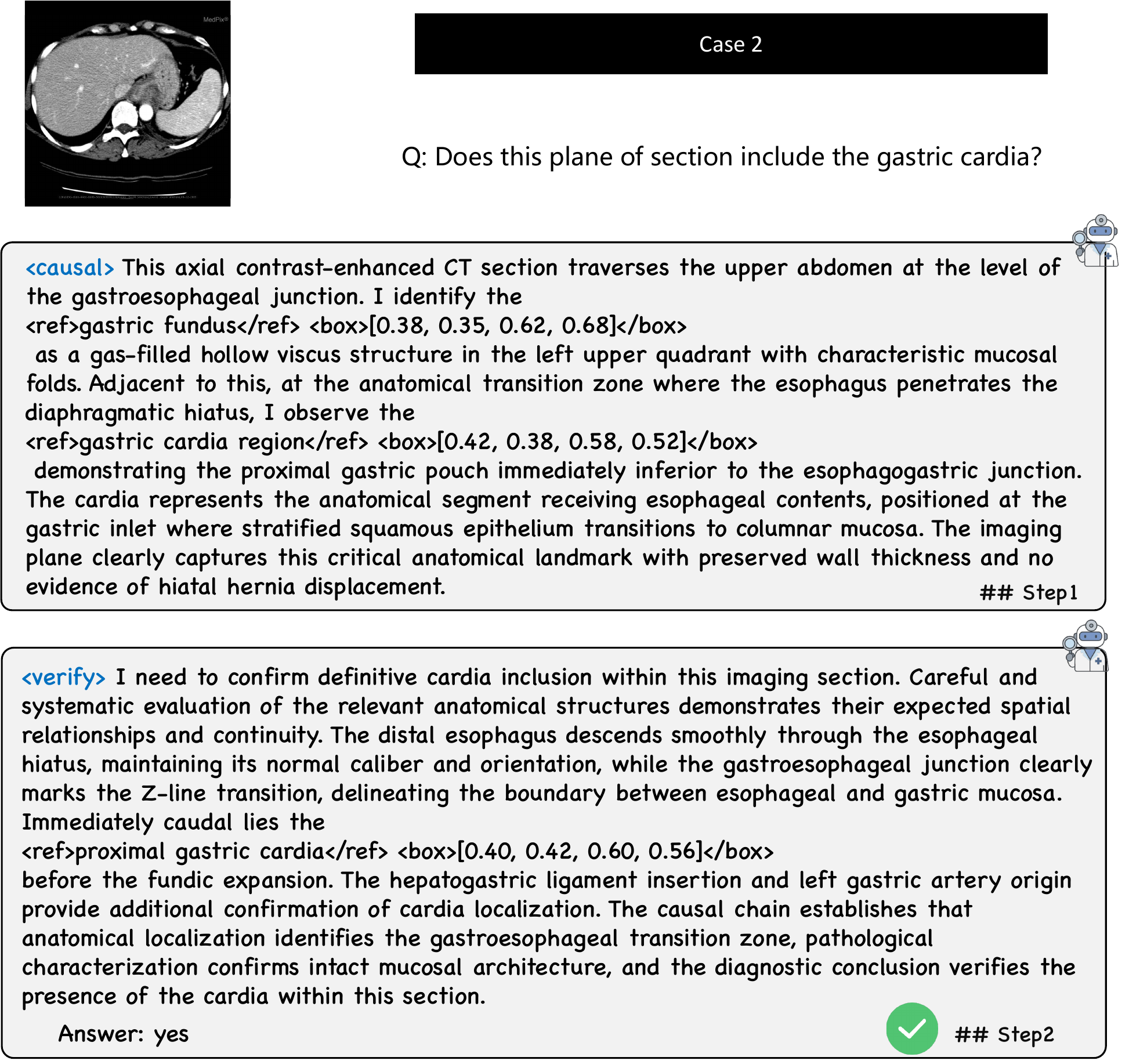}
    \caption{Abdominal examination section validating gastric cardia identification through causal anatomical landmark detection, confirming presence via systematic spatial relationship analysis with esophagogastric junction.}
    \label{fig:case2_cardia}
\end{figure*}

\begin{figure*}[t]
    \centering
    \includegraphics[width=0.8\textwidth]{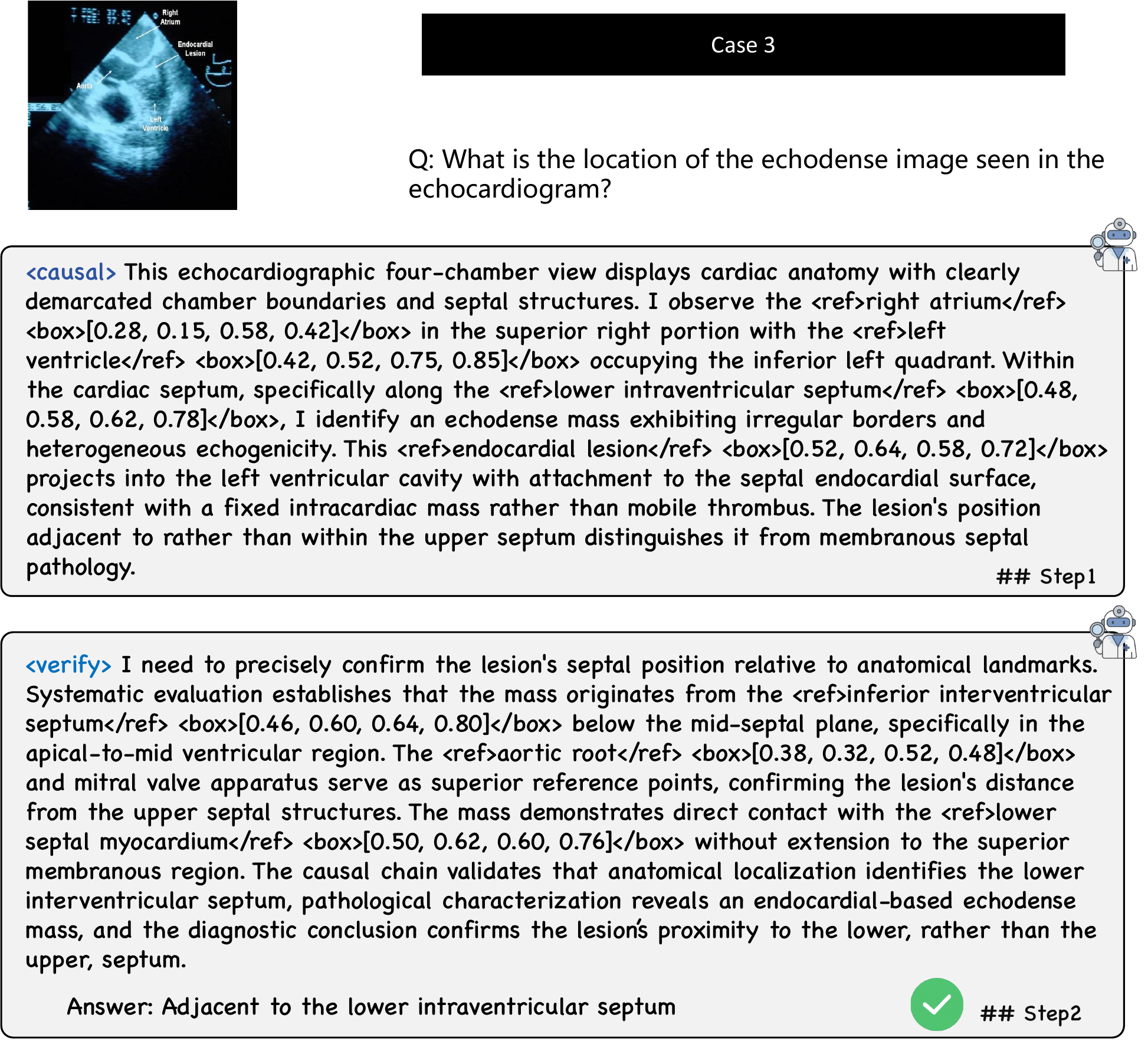}
    \caption{Echocardiographic image demonstrating successful causal localization of endocardial lesion positioned adjacent to lower intraventricular septum through structured anatomical-pathological reasoning chain.}
    \label{fig:case3_echo}
\end{figure*}

\begin{figure*}[t]
    \centering
    \includegraphics[width=0.8\textwidth]{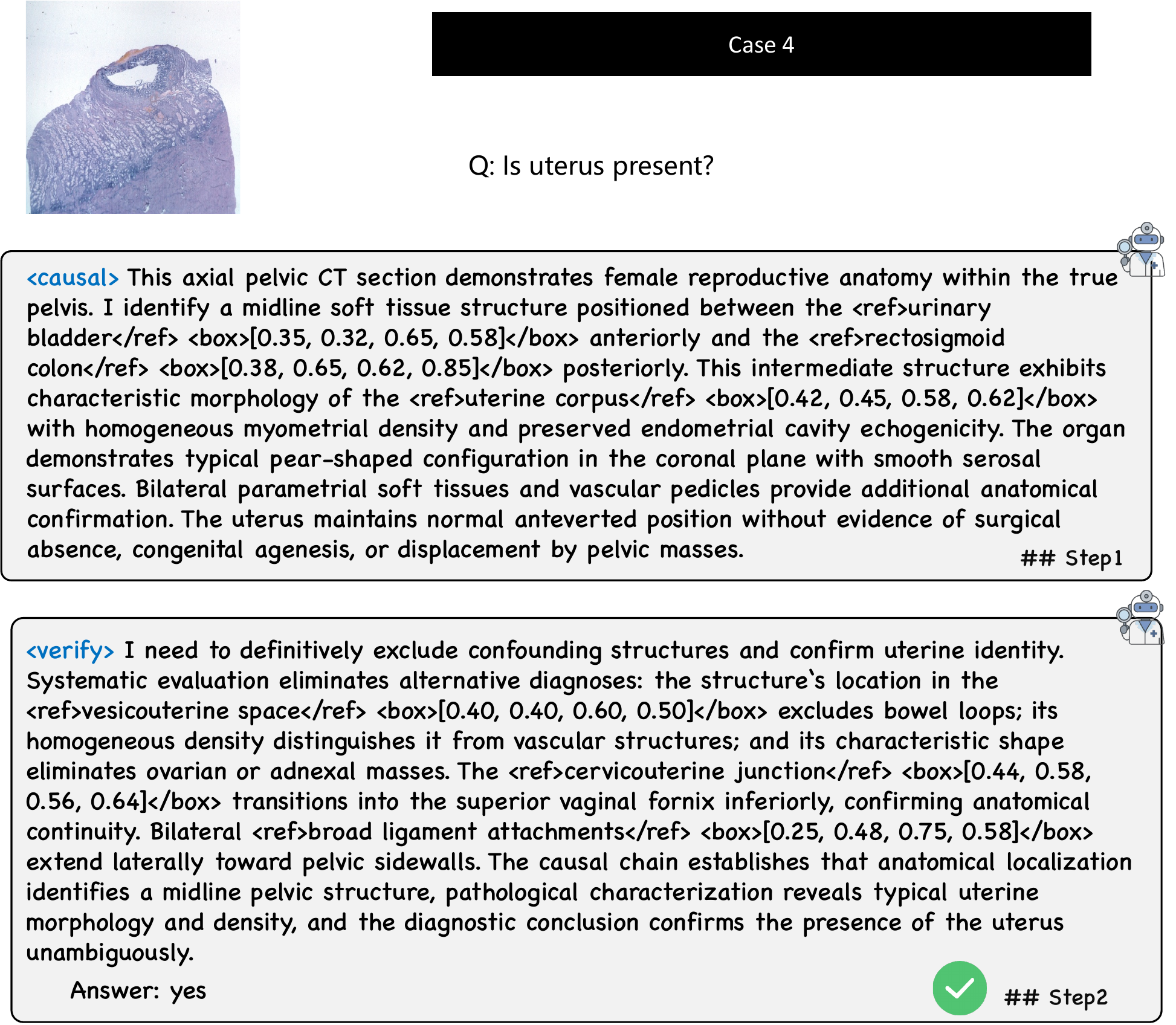}
    \caption{Pelvic CT scan validating binary anatomical detection where MedCausalX correctly confirms uterine presence through causal verification eliminating confounding structures.}
    \label{fig:case4_uterus}
\end{figure*}

MedCausalX demonstrates robust causal reasoning across multiple public medical VQA datasets. In Figure~\ref{fig:case1_lung} (Case 1), the model successfully performs multi-object anatomical quantification, accurately identifying bilateral pulmonary structures via causal factorization. Figure~\ref{fig:case2_cardia} (Case 2) illustrates its capability in anatomical landmark detection, where the presence of the gastric cardia is correctly confirmed through systematic spatial reasoning. In Figure~\ref{fig:case3_echo} (Case 3), MedCausalX shows precise spatial localization by identifying an endocardial lesion adjacent to the lower intraventricular septum, leveraging structured anatomical factorization ($A \rightarrow P \rightarrow Y$). Finally, Figure~\ref{fig:case4_uterus} (Case 4) demonstrates binary anatomical detection, with the model reliably confirming uterine presence through causal verification. Together, these examples highlight that MedCausalX grounds its diagnostic predictions in genuine anatomical evidence, rather than relying on superficial pattern correlations.

\subsection{Prompt Engineering and Reflective Tokens}

\begin{figure*}[t]
    \centering
    \includegraphics[width=0.9\textwidth]{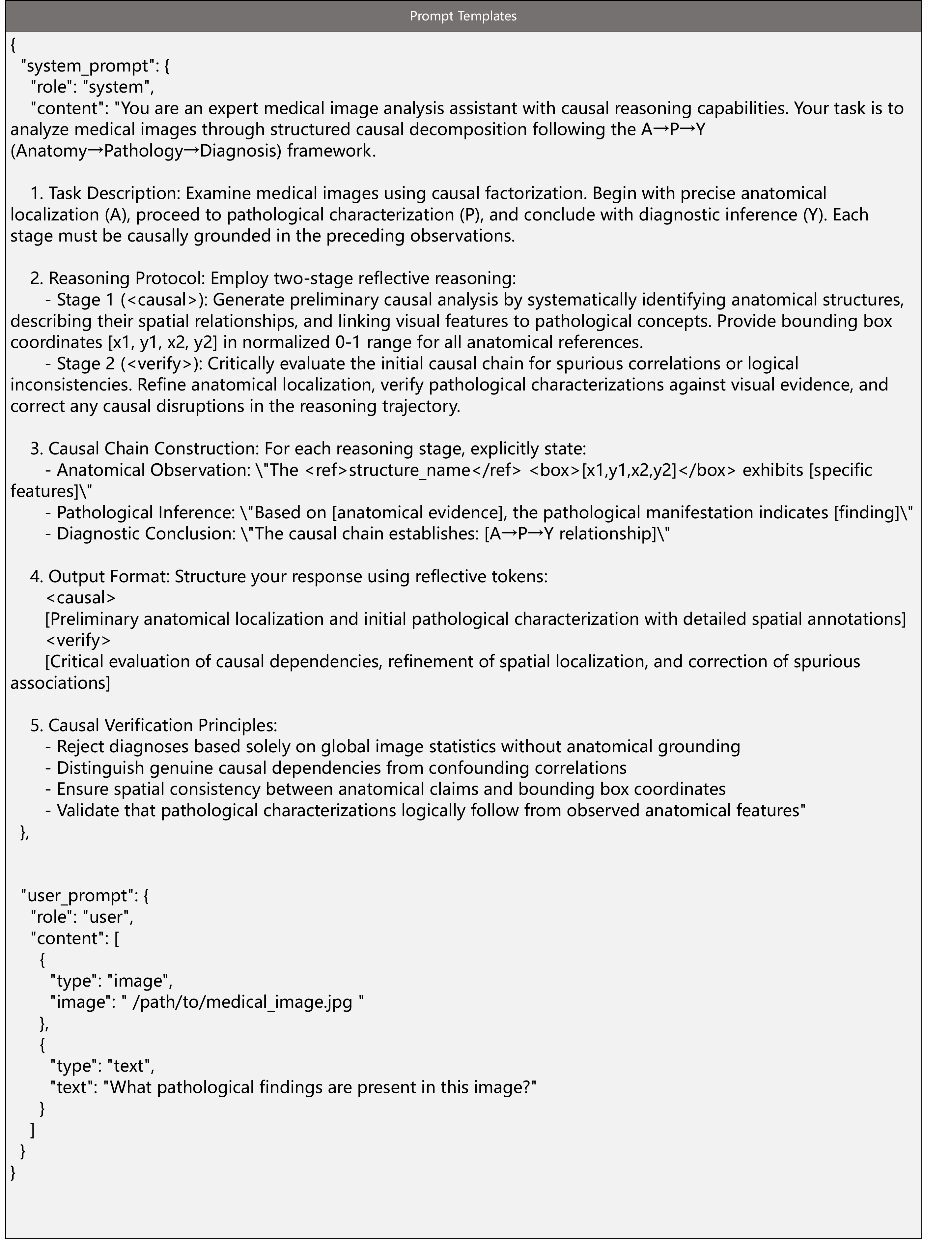}
    \caption{Structured prompt template for MedCausalX showing system instructions for pathological analysis approach and output format specifications enabling causal chain generation.}
    \label{fig:prompt_system}
\end{figure*}

MedCausalX utilizes a structured prompt template (Figure~\ref{fig:prompt_system}) to guide causal reasoning through explicit task decomposition. The system prompt directs the model to systematically examine medical images, compare findings with initial predictions, and link observations to the provided concept vocabulary using concrete visual evidence. This hierarchical framework ensures semantic consistency with medical ontologies while producing declarative sentences suitable for causal verification. Reflective tokens \texttt{<causal>} and \texttt{<verify>} facilitate dynamic reasoning mode transitions, allowing preliminary anatomical analysis to trigger subsequent verification and correction phases whenever inconsistencies are detected.

\begin{table}[t]
\centering
\renewcommand{\arraystretch}{1.08}
\caption{Ablation study on training strategy configuration. Light blue background indicates selected configuration, bold values denote best performance in each row.}
\label{tab:training_strategy_ablation}
\resizebox{\columnwidth}{!}{
\begin{tabular}{cc|cccc|c}
\toprule[1pt]
\hline
\textbf{Config.} & \textbf{Value} & \textbf{VQA-RAD} & \textbf{SLAKE} & \textbf{PathVQA} & \textbf{PMC-VQA} & \textbf{Average} \\
\midrule[0.6pt]
\multicolumn{7}{l}{\textit{(a) Training Stage Order}} \\
\midrule[0.6pt]
\multirow{2}{*}{\rotatebox{0}{\textbf{Seq.}}}
& GRPO→DPO & 73.5 & 81.4 & 68.2 & 78.9 & 75.5 \\
& \cellcolor[RGB]{231, 248, 254}\textbf{DPO→GRPO}
& \cellcolor[RGB]{231, 248, 254}\textbf{79.8}
& \cellcolor[RGB]{231, 248, 254}\textbf{87.2}
& \cellcolor[RGB]{231, 248, 254}\textbf{73.2}
& \cellcolor[RGB]{231, 248, 254}\textbf{84.6}
& \cellcolor[RGB]{231, 248, 254}\textbf{81.2} \\
\midrule[0.6pt]
\multicolumn{6}{l}{\textit{(b) GRPO Group Size $G$}} \\
\midrule[0.6pt]
\multirow{5}{*}{\rotatebox{0}{\textbf{$G$}}}
& 4 & 76.8 & 84.3 & 70.5 & 81.7 & 78.3 \\
& 6 & 78.5 & 86.1 & 72.1 & 83.5 & 80.1 \\
& \cellcolor[RGB]{231, 248, 254}\textbf{8}
& \cellcolor[RGB]{231, 248, 254}\textbf{79.8}
& \cellcolor[RGB]{231, 248, 254}\text{87.2}
& \cellcolor[RGB]{231, 248, 254}\textbf{73.2}
& \cellcolor[RGB]{231, 248, 254}\textbf{84.6}
& \cellcolor[RGB]{231, 248, 254}\textbf{81.2} \\
& 12 & 78.9 & \textbf{88.5} & 72.6 & 83.9 & 80.9 \\
& 16 & 77.4 & 85.2 & 71.3 & 82.4 & 79.1 \\
\hline
\bottomrule[1pt]
\end{tabular}
}
\end{table}

\subsection{Efficiency-Performance Analysis}

Figure~\ref{fig:efficiency} compares efficiency and performance of state-of-the-art medical vision-language models using a composite score averaging diagnostic accuracy and spatial grounding IoU. General-purpose VLMs improve modestly with scale but remain below medical-specific models. Domain-adapted medical VLMs achieve higher performance, and chain-of-thought enhanced models further benefit from structured reasoning. MedCausalX sets a new state-of-the-art, outperforming comparable 32B models and even larger specialized systems while using fewer parameters. These results highlight that explicit causal reasoning and structured factorization provide substantial gains beyond raw parameter scaling, establishing MedCausalX as the most efficient and effective solution in its class.

\subsection{MedCausalX Scaling Analysis}

%Table~\ref{tab:model_scaling} examines how model capacity influences causal reasoning across region-centric medical tasks. The 3B model captures basic anatomical structures (53.42\% BLEU-1, 68.45\% Object-F1) but struggles with lesion-level reasoning. Expanding to 14B parameters yields significant gains (+20.07 BLEU-1, +17.68 Acc), indicating enhanced causal decomposition and hierarchical understanding. Scaling further to 32B achieves the best overall performance (79.82\% BLEU-1, 46.82\% Region-F1), though improvements diminish beyond 14B. Notably, the 32B variant attains 88.45\% BertScore, reflecting superior semantic coherence. These results confirm that larger models enable deeper causal verification and stronger spatial–linguistic alignment, validating MedCausalX’s scalable architecture for complex medical reasoning.

Table~\ref{tab:model_scaling} examines how model capacity influences causal reasoning. The 3B variant captures basic anatomical structures but shows clear limitations in lesion-level reasoning, achieving only 53.42\% BLEU-1 for structure identification. Scaling to 14B parameters improves performance by 20 approximately points, indicating better causal decomposition and  understanding of medical concepts. The 32B variant achieves the best overall results, with BLEU-1 of 79.82\% and BertScore of 88.45\%, demonstrating strong semantic coherence, precise spatial grounding, and consistent reasoning quality. These trends confirm that larger models enable deeper causal verification and stronger spatial–linguistic alignment, supporting MedCausalX’s scalable architecture for complex medical reasoning.

\subsection{Model Architecture Effect}

Table~\ref{tab:model_scale_ablation} highlights how architectural design and model capacity affect VQA reasoning. MedCausalX-32B achieves 81.2\% average accuracy, outperforming CoT baselines Med-R1 and MedVLM-R1 by 2–3 points, demonstrating that its causal factorization and error-attributed reinforcement learning architecture enables structured reasoning beyond standard chain-of-thought. Performance scales from 68.6\% at 3B to 81.2\% at 32B, showing that counterfactual reasoning benefits from increased capacity. General VLMs LLaVA-1.5 (55.2\%) and InternVL-2 (63.4\%) lag behind, emphasizing the importance of specialized architecture for anatomically grounded diagnostic tasks.

\subsection{Training Strategy Ablation}

Table~\ref{tab:training_strategy_ablation} validates our dual-stage reinforcement learning design. The DPO→GRPO sequence achieves 81.2\% average accuracy, outperforming the reversed order by 5.7 points. This indicates that applying DPO first allows off-policy learning from error trajectories, effectively establishing robust reasoning foundations, while subsequent GRPO exploration refines causal intervention strategies without premature convergence to superficial patterns. Regarding group size configuration, $G$=8 provides an optimal balance: smaller groups introduce excessive gradient noise that destabilizes causal learning, whereas larger configurations, such as $G$=12 or 16, exhibit diminishing returns because conservative advantage normalization suppresses the discovery of novel interventions, which is essential for adaptive and flexible reasoning.

\subsection{Reward Component Configuration}

\begin{table}[t]
\centering
\renewcommand{\arraystretch}{1.08}
\caption{Ablation of reward components and error localization threshold.  Light blue background indicates selected configuration.}
\label{tab:reward_ablation}
\resizebox{\columnwidth}{!}{
\begin{tabular}{cc|cccc|c}
\toprule[1pt]
\hline
\textbf{Config.} & \textbf{Value} & \textbf{VQA-RAD} & \textbf{SLAKE} & \textbf{PathVQA} & \textbf{PMC-VQA} & \textbf{Average} \\
\midrule[0.6pt]
\multicolumn{7}{l}{\textit{(a) Reward Weight Configuration $\lambda_{\text{acc}}/\lambda_{\text{causal}}$}} \\
\midrule[0.6pt]
\multirow{5}{*}{\rotatebox{0}{\textbf{$\lambda$}}}
& 1.0/0.0 & 71.5 & 78.9 & 65.3 & 76.2 & 73.0 \\
& 0.7/0.3 & 76.8 & 84.1 & 70.5 & 81.7 & 78.3 \\
& \cellcolor[RGB]{231, 248, 254}\textbf{0.5/0.5}
& \cellcolor[RGB]{231, 248, 254}\textbf{79.8}
& \cellcolor[RGB]{231, 248, 254}\textbf{87.2}
& \cellcolor[RGB]{231, 248, 254}\textbf{73.2}
& \cellcolor[RGB]{231, 248, 254}\textbf{84.6}
& \cellcolor[RGB]{231, 248, 254}\textbf{81.2} \\
& 0.3/0.7 & 78.2 & 85.7 & 71.8 & 83.1 & 79.7 \\
& 0.0/1.0 & 67.3 & 74.5 & 61.2 & 71.8 & 68.7 \\
\midrule[0.6pt]
\multicolumn{7}{l}{\textit{(b) Error Localization Threshold $\tau$}} \\
\midrule[0.6pt]
\multirow{5}{*}{\rotatebox{0}{\textbf{$\tau$}}}
& 0.3 & 70.8 & 77.5 & 64.1 & 75.3 & 71.9 \\
%& 0.5 & 70.3 & 78.8 & 64.9 & 69.8 & 71.0 \\
& 0.5 & 75.3 & 82.9 & 69.4 & 80.5 & 77.0 \\
& \cellcolor[RGB]{231, 248, 254}\textbf{0.7}
& \cellcolor[RGB]{231, 248, 254}\textbf{79.8}
& \cellcolor[RGB]{231, 248, 254}\textbf{87.2}
& \cellcolor[RGB]{231, 248, 254}\textbf{73.2}
& \cellcolor[RGB]{231, 248, 254}\textbf{84.6}
& \cellcolor[RGB]{231, 248, 254}\textbf{81.2} \\
%& 0.75 & 71.8 & 79.7 & 65.8 & 71.1 & 72.1 \\
& 0.8 & 77.1 & 84.6 & 70.8 & 82.3 & 78.7 \\
& 0.9 & 72.4 & 79.8 & 66.5 & 77.1 & 73.9 \\
\hline
\bottomrule[1pt]
\end{tabular}
}
\end{table}

\begin{table}[t]
\centering
\renewcommand{\arraystretch}{1.08}
\caption{Out-of-domain evaluation on specialized medical VQA datasets. Models trained on core benchmarks and evaluated zero-shot on unseen domains.}
\label{tab:cross_dataset_generalization}
\resizebox{\columnwidth}{!}{
\begin{tabular}{l|ccc|c}
\toprule[1pt]
\hline
\textbf{Method} & \textbf{OVQA} & \textbf{EndoVis-VQA} & \textbf{Skin-VQA} & \textbf{Average} \\
\midrule[0.6pt]
GPT-4o~\cite{hurst2024gpt} & 43.5 & 39.8 & 46.1 & 43.1 \\
LLaVA-Med~\cite{li2023llava} & 47.2 & 44.3 & 49.5 & 47.0 \\
InternVL-2~\cite{chen2024expanding} & 49.8 & 46.7 & 51.2 & 49.2 \\
Med-R1~\cite{lai2026med} & 44.8 & 40.5 & 46.3 & 43.9 \\
MedVLM-R1~\cite{pan2025medvlm} & 50.3 & 47.1 & 49.8 & 49.1 \\
\midrule[0.6pt]
\cellcolor[RGB]{231, 248, 254}\textbf{MedCausalX (Ours)}
& \cellcolor[RGB]{231, 248, 254}\textbf{58.4}
& \cellcolor[RGB]{231, 248, 254}\textbf{54.2}
& \cellcolor[RGB]{231, 248, 254}\textbf{59.7}
& \cellcolor[RGB]{231, 248, 254}\textbf{57.4} \\
\hline
\bottomrule[1pt]
\end{tabular}
}
\end{table}

Table~\ref{tab:reward_ablation} highlights the interdependence of reward components in causal reasoning optimization. Balanced weighting achieves the best accuracy, outperforming accuracy-only and causal-only setups, showing that multi-objective optimization prevents overfitting to diagnostic shortcuts while preserving reasoning quality. For error localization threshold $\tau$, our study shows that 0.7 provides the best trade-off between sensitivity and specificity: lower thresholds generate excessive false positives, while higher thresholds miss critical reasoning errors, confirming that precise error attribution is essential for effective causal training.

\subsection{Cross-Dataset Generalization}

\begin{figure}[t]
    \centering
    \includegraphics[width=\linewidth]{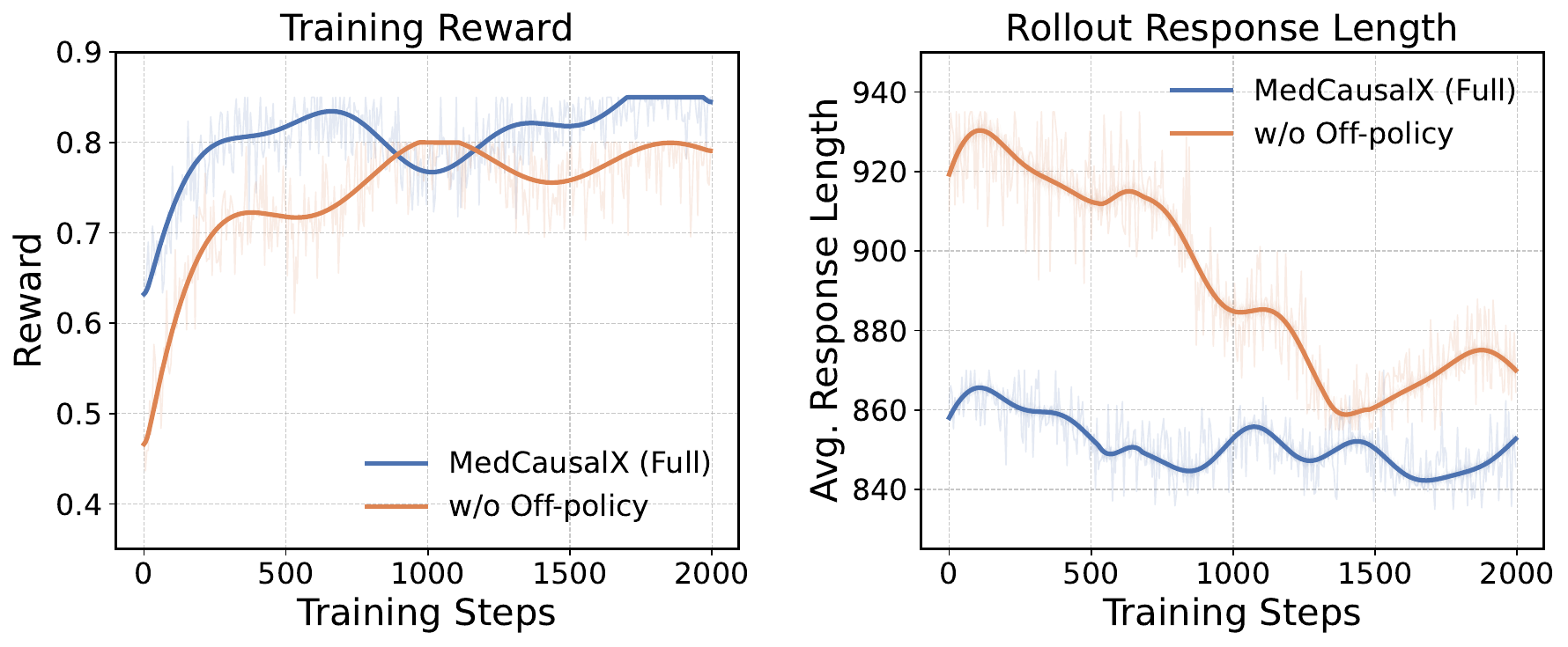}
\caption{Training dynamics of on-policy GRPO optimization. MedCausalX (Full) with DPO pretraining achieves higher rewards and maintains stable, compact reasoning compared to direct GRPO training from preliminary SFT baseline.}   \label{fig:training_dynamics_supp}
\end{figure}

%, outperforming the strongest baseline, MedVLM-R1, by 8.3 points
Table~\ref{tab:cross_dataset_generalization} demonstrates MedCausalX's superior zero-shot transfer across specialized medical domains. The model achieves an average accuracy of 57.4\%. Gains are particularly notable on domain-specific tasks, including OVQA in ophthalmology with an improvement of 8.1\% and Skin VQA in dermatology with an improvement of 9.9\%. These results suggest that causal reasoning mechanisms, such as anatomical localization, pathology identification, and counterfactual verification, provide structural inductive biases that generalize effectively across modalities and anatomical regions. In contrast, reasoning-enhanced methods like Med-R1 and MedVLM-R1 achieve only modest improvements over standard models such as InternVL-2, indicating that chain-of-thought approaches lack sufficient compositional structure for reliable domain generalization.

\subsection{Training Dynamics Analysis}
Figure~\ref{fig:training_dynamics_supp} illustrates MedCausalX's training dynamics under on-policy optimization. The reward curves show steady improvement for both configurations, with MedCausalX (Full) benefiting from off-policy initialization to start higher. While the w/o Off-policy baseline gradually catches up, the Full version maintains a consistent advantage, reaching 0.845 versus 0.791. Response lengths reveal distinct trajectories: Full maintains compact reasoning around 850 tokens, reflecting DPO's pre-established efficiency. In contrast, the w/o Off-policy baseline starts at 919 tokens and stabilizes around 870 tokens as GRPO converges. These patterns confirm that off-policy training provides immediate efficiency gains, while GRPO alone requires longer optimization to achieve comparable compactness.
%Figure~\ref{fig:training_dynamics_supp} illustrates the training dynamics of MedCausalX during on-policy GRPO optimization. The reward curves show steady improvement for both configurations, with MedCausalX (Full) benefiting from off-policy initialization to start higher. While the w/o Off-policy baseline gradually catches up, the Full version maintains a consistent advantage, reaching 0.845 versus 0.791, a 5.4-point gap. Rollout response lengths reveal distinct trajectories: MedCausalX (Full) maintains stable, compact reasoning around 850 tokens throughout training, reflecting DPO's pre-established efficiency. In contrast, the w/o Off-policy baseline exhibits a pronounced learning curve, starting at 919 tokens and ultimately stabilizing around 870 tokens as GRPO converges. These patterns confirm that off-policy training provides immediate efficiency benefits, while direct GRPO requires longer optimization to achieve comparable compactness.
\clearpage

\clearpage
\newpage

{
    \small
    \bibliographystyle{ieeenat_fullname}
    \bibliography{main}
}


\end{document}